%% file: LRQ_arxiv.tex
\documentclass{article}

\PassOptionsToPackage{numbers,compress}{natbib}

\usepackage[preprint]{submission}

\usepackage[utf8]{inputenc} 
\usepackage[T1]{fontenc}    
\usepackage{hyperref}       
\usepackage{url}            
\usepackage{booktabs}       
\usepackage{amsfonts}       
\usepackage{float} 
\usepackage{nicefrac}       
\usepackage{microtype}      
\usepackage{xcolor}         
\usepackage{bm}
\usepackage{graphicx}
\usepackage{amsmath}
\usepackage{amssymb}
\usepackage{mathtools}
\usepackage{amsthm}
\usepackage{multirow}
\usepackage{hyperref}
\usepackage{algorithm}
\usepackage{algorithmicx}
\usepackage{algpseudocode}
\usepackage{adjustbox}
\usepackage{wrapfig}
\usepackage{pifont}
\usepackage{enumitem}
\usepackage{amsthm}
\newtheorem{lemma}{Lemma}
\newtheorem{theorem}{Theorem}
\usepackage{makecell}

\title{LRQ-DiT: Log-Rotation Post-Training Quantization \\ of Diffusion Transformers \\ for Image and Video Generation}

\author{%
  Lianwei Yang\footnotemark[1] $^{\ 1,2}$, ~~
  Haokun Lin\footnotemark[1] $^{\ 1,2,5}$, ~~
  Tianchen Zhao\footnotemark[1] $^{\ 3}$, ~~
  Yichen Wu $^{\ 4,5}$, ~\\
  \textbf{Hongyu Zhu $^{\ 3}$}, ~
  \textbf{Ruiqi Xie $^{\ 3}$}, ~
  \textbf{Zhenan Sun $^{\ 1,2}$}, ~
  \textbf{Yu Wang $^{\ 3}$}, ~
  \textbf{Qingyi Gu\footnotemark[2] $^{\ 1}$} ~ \\
  $^{1}$ Institute of Automation, Chinese Academy of Sciences \\ 
  $^{2}$ School of Artificial Intelligence, University of Chinese Academy of Sciences \\
  $^{3}$ Department of Electronic Engineering, Tsinghua University \\
  $^{4}$  School of Engineering and Applied Sciences, Harvard University \\
  $^{5}$  Department of Computer Science, City University of Hong Kong \\
}

\begin{document}
\renewcommand{\thefootnote}{\fnsymbol{footnote}}
\footnotetext[1]{Equal contribution.}  
\footnotetext[2]{Corresponding authors.}
\vspace{-0.6cm}
\maketitle

\input{Sec/0_abs}
\input{Sec/1_intro}

\input{Sec/2_related}
\input{Sec/3_preliminaries}

\input{Sec/4_method}

\input{Sec/5_exp}

\input{Sec/6_conclusion}

\newpage
\bibliography{reference}
\bibliographystyle{plainnat}

\newpage
\input{Sec/7_appendix}

\end{document}

%% file: Sec/0_abs.tex
\begin{abstract}
Diffusion Transformers (DiTs) have achieved impressive performance in text-to-image and text-to-video generation.
However, their high computational cost and large parameter sizes pose significant challenges for usage in resource-constrained scenarios.
Effective compression of models has become a crucial issue that urgently needs to be addressed.
Post-training quantization (PTQ) is a promising solution to reduce memory usage and accelerate inference, but existing PTQ methods suffer from severe performance degradation under extreme low-bit settings.
After experiments and analysis, we identify two key obstacles to low-bit PTQ for DiTs:
\textbf{(1)} the weights of DiT models follow a Gaussian-like distribution with long tails, causing uniform quantization to poorly allocate intervals and leading to significant quantization errors. 
This issue has been observed in the linear layer weights of different DiT models, which deeply limits the performance.
\textbf{(2)} two types of activation outliers in DiT models: (i) Mild Outliers with slightly elevated values, and (ii) Salient Outliers with large magnitudes concentrated in specific channels, which disrupt activation quantization.
To address these issues, we propose \textbf{LRQ-DiT}, an efficient and accurate post-training quantization framework for image and video generation.
First, we introduce \textbf{T}win-\textbf{L}og \textbf{Q}uantization (TLQ), 
a log-based method that allocates more quantization intervals to the intermediate dense regions, effectively achieving alignment with the weight distribution and reducing quantization errors.
Second, we propose an \textbf{A}daptive \textbf{R}otation \textbf{S}cheme (ARS) that dynamically applies Hadamard or outlier-aware rotations based on activation fluctuation, effectively mitigating the impact of both types of outliers.
Extensive experiments on various text-to-image and text-to-video DiT models demonstrate that LRQ-DiT preserves high generation quality under low-bit weight–activation quantization, outperforming existing PTQ baselines.
\end{abstract}

%% file: Sec/1_intro.tex
\section{Introduction}
\begin{figure*}[t]
\centering
\includegraphics[width=1\textwidth]{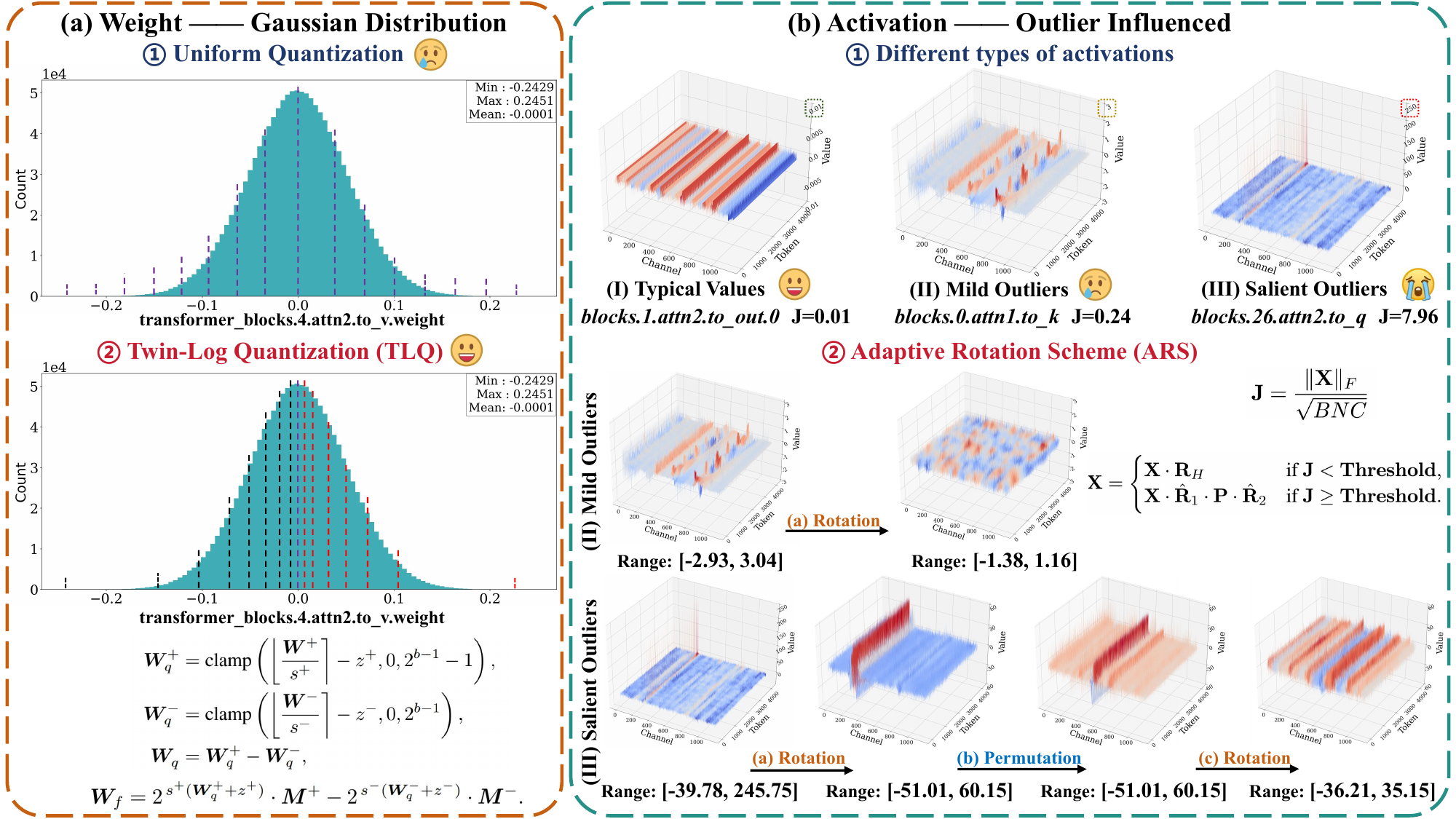} 
\caption{The overall framework of LRQ-DiT. 
(a) LRQ-DiT proposes the twin-log quantization (TLQ) method to address imbalanced weight distribution and excessive quantization errors. Compared to uniform quantization, TLQ significantly enhances the practicality of low-bit quantization.
(b) LRQ-DiT utilizes the metric \textbf{J} to identify mild and salient outliers in activations, and proposes an adaptive rotation scheme (ARS) that applies customized rotations to suppress the impact of each type of outlier in activations.
}
\vskip 0.1in
\label{fig:overview}
\end{figure*}

Diffusion models have achieved remarkable success in text-to-image and text-to-video generation tasks, with recent advances expanding their applications across vision domains~\cite{esser2024scalingSD3,podell2023sdxl,xie2024sana, ying2025diff, liu2024diffvein, lin2025toklip, jiangimage, zhou2024doge}. 
Notably, diffusion transformers (DiTs) replace the traditional U-Net backbone with the Transformer architecture, significantly improving scalability and computational efficiency.  
DiTs are becoming dominant in tasks such as image generation~\cite{ chen2025conceptcraft}, video generation~\cite{zhang2025tvg}, and image super-resolution~\cite{korkmaz2025leveraging, zhang2024spatial}. 
Despite their performance, the large number of parameters in DiTs poses serious challenges for deployment on resource-constrained edge devices, especially in terms of memory consumption and inference latency.

Model compression~\cite{xu2022improving, zhao2023exploiting, song2025exploring, zhang2025dfmc, lin2024mope} offers a promising acceleration solution, with quantization being one of the most effective techniques.
By converting high-precision floating-point values into low-bit integers, quantization reduces both storage and computation costs, and has been widely adopted in large-scale Transformers.
Compared with quantization-aware training (QAT)~\cite{ guo2025shiftquant, xiao2024binaryvit},
PTQ has become increasingly popular for large models, such as Large Language Models (LLMs)~\cite{yao2022zeroquant,shao2023omniquant} and Diffusion Transformers (DiTs), due to its training-free nature.  
Existing PTQ methods for DiTs typically focus on moderate configurations such as W4A8 and W8A8, but performance under lower-bit settings remains limited.  
PTQ4DiT~\cite{wu2024ptq4dit} and Q-DiT~\cite{chen2025qdit} incorporate time-step-specific activation patterns to achieve W4A8 quantization on DiT-XL/2, yet they exhibit performance degradation when applied to more advanced DiT variants such as PixArt and FLUX. 
ViDiT-Q~\cite{zhao2024vidit} proposes a dynamic quantization strategy, and SVDQuant~\cite{li2024svdquant} addresses outliers via kernel-level protection, both 
improving the performance of W4A8 quantized models.
Besides, some studies~\cite{feng2025q, li2025dvd} have made preliminary attempts on video generation models.
However, achieving accurate quantization for DiTs remains a significant challenge, particularly due to the difficulty in preserving precision for both weights and activations under extreme low-bit constraints.

Our preliminary experiments identify two key challenges for low-bit quantization of DiT models: 
(1) \textit{Uniform quantizers fundamentally mismatch the Gaussian-like weight distribution of DiTs.}
We observe that almost all weight matrices in DiT models follow a Gaussian-like distribution with long tails on both sides.
The Uniform quantization fails to allocate sufficient quantization intervals to the dense central region and ignores the long-tail extremes, resulting in large quantization errors under low-bit settings.
(2) \textit{DiT activations contain two distinct types of outliers that disrupt low-bit quantization.}
We discover that DiT models exhibit two distinct types of outliers.
The majority of outliers in activations are Mild Outliers, with values slightly exceeding normal activation levels and rarely surpassing 5.
A smaller portion is Salient Outliers, represented as high-magnitude activations concentrated in specific channels with their values approaching 300.
These outliers pose significant challenges for low-bit activation quantization.

To address the challenges identified above, we propose an accurate and efficient method \textbf{LRQ-DiT}, which integrates two core components: a twin-log quantization (TLQ) method for Gaussian-like model weights, and an adaptive rotation scheme (ARS) for accurate activations quantization.
First, to better match the Gaussian distribution of weights, we introduce \textbf{twin-log quantization}.
We apply logarithmic transformation to the model weights and quantize the positive and negative parts separately, allocating more quantization intervals to the dense areas in the middle.
A simple search-based clipping strategy is further employed to suppress extreme values in the long-tail regions.
Second, we propose an \textbf{adaptive rotation scheme} to handle Mild and Salient activation outliers.
To balance the accuracy and efficiency, we define an adaptive metric to measure the fluctuation level of activations across layers.
Layers with low fluctuation are processed using Hadamard rotation, while layers with high fluctuation adopt dual transformations, combining greedy outlier-aware rotation with channel-wise permutation for enhanced precision.
This design enables LRQ-DiT to effectively mitigate both types of outliers,
while maintaining high precision and deployment efficiency. 
Finally, comprehensive evaluations on PixArt and FLUX models for text-to-image generation, as well as OpenSORA model for text-to-video generation, prove the superiority of LRQ-DiT in achieving low-bit quantization without sacrificing generation quality.
Our main contributions are summarized as follows:
\begin{itemize}[leftmargin=4mm]
\item We observe that uniform quantization struggles to adapt to the Gaussian-like distribution of weights, leading to significant quantization errors. To address this issue, we propose twin-log quantization to resolve this issue.
\item We leverage the metric to identify both mild and salient outliers in activations and propose an adaptive rotation scheme to perform targeted rotations for these two types of outliers, achieving effective outlier suppression.
\item Extensive quantization experiments conducted on PixArt, FLUX and OpenSORA models under varying bit settings and different prompt sets demonstrate the effectiveness and robustness of LRQ-DiT, especially in low-bit scenarios.
\end{itemize}

%% file: Sec/2_related.tex
\section{Related Works}
\subsection{Text-to-Image Models}
Text-to-image generation has made remarkable progress, largely driven by advancements in model architecture.
Stable Diffusion~\citep{rombach2022high} bridges the semantic gap between text and image modalities, and its successor, Stable Diffusion 3~\citep{esser2024scalingSD3}, further enhances multi-modal fusion.
Subsequent models such as SDXL~\citep{podell2023sdxl} and its variants provide a strong foundation for efficient and high-quality text-to-image generation.
More recently, transformer-based architectures have replaced U-Net in diffusion models, opening new possibilities for downstream applications.
PixArt-$\alpha$~\citep{chen2023pixartalpha} reduces training cost while maintaining image quality, and PixArt-$\Sigma$~\citep{chen2024pixartsigma} significantly improves 4K image generation efficiency.
FLUX~\citep{flux} (primarily FLUX.1-schnell and FLUX.1-dev) further demonstrates the capability of DiT-based models to generate high-resolution, high-fidelity images.
OpenSORA~\cite{zheng2024open} brings DiTs to the video domain with a real-time capable, multimodal generation pipeline.
However, the parameter and computational complexity of the model are constantly increasing, which poses difficulties for further development and deployment.

\subsection{Text-to-Video Models}
Given that DiT has achieved remarkable success and garnered significant attention in text-to-image generation tasks, numerous studies have attempted to introduce it into text-to-video generation tasks to further expand its range of applications.
The Latte~\citep{ma2024latte} model achieves efficient and high-quality video generation through its latent diffusion transformer architecture.
Sora~\citep{liu2024sora} has been trained for generating clear and realistic scene videos from textual instructions.
OpenSORA~\citep{zheng2024open} employs the Spatial-Temporal Diffusion Transformer~(STDiT) technology to accomplish a variety of visual generation tasks, facilitating the synthesis of high-definition videos.
CogVideoX~\citep{yang2024cogvideox} employs progressive training with multi-resolution frame packing to generate high-quality, coherent, and long-duration videos.
HunyuanVideo~\citep{kong2024hunyuanvideo} trains a video generation model with over 13 billion parameters.
Wan~\citep{wan2025wan} has made substantial progress in generation by introducing innovative methods based on the DiT framework.
These works have greatly expanded the applications of DiT models and achieved outstanding results.
However, as the number of model parameters and computational complexity continue to increase, it also presents significant challenges for further applications.
This issue is particularly evident in the text-to-video generation task.

\subsection{Model Quantization}

Quantization reduces model memory by converting high-precision floating-point weights to low-bit integers, while activation quantization accelerates inference. 
\textbf{For Vision Transformers (ViTs)}, several studies have proposed more accurate quantizers, such as twin uniform quantizer~\citep{yuan2022ptq4vit}, log2 quantizer~\citep{lin2022fqvit}, log$\sqrt{2}$ quantizer~\citep{Repq-vit}, shift-uniform-log2 quantizer~\citep{zhongyunshan} and other methods~\citep{yang2024mgrq, yang2024dopq, xiao2023patch, xiao2024ttaq}. 
These quantizers are designed to match the characteristics of weights or activations, achieving excellent performance.
Post-training quantization (PTQ) is also widely used in large-scale pre-trained models as it avoids costly retraining. 
\textbf{For Large Language Models (LLMs)}, outliers~\citep{dettmers2022llm_int8,sun2024massive}, the values significantly larger than most activations, pose a major challenge. 
SmoothQuant~\citep{xiao2023smoothquant} shifts quantization difficulty to weights for better 8-bit performance, while rotation-based methods such as QuaRot~\citep{ashkboos2024quarot} and DuQuant~\citep{lin2024duquant} balance outliers via rotation transformations, enabling competitive 4-bit weight–activation quantization.
Applying these techniques to diffusion models is nontrivial due to time-step-dependent features. 
\textbf{For Diffusion Models (DMs)}, Q-Diffusion~\citep{li2023q} and PTQ4DM~\citep{shang2023post_PTQ4DM} considered the influence of time-steps when determining quantization parameters, achieving effective 8-bit quantization, which also stimulated subsequent works~\citep{he2023ptqd, huang2024tfmq, wang2024quest}.
However, there are still differences in the structure between DiTs and DMs.
\textbf{For Diffusion Transformers (DiTs)}, to address these variations, Q-DiT~\citep{chen2025qdit} assigns channel-wise quantization parameters, PTQ4DiT~\citep{wu2024ptq4dit} reallocates salient channels across time steps, SVDQuant~\citep{li2024svdquant} protects activation outliers with low-rank branches, and ViDiT-Q~\citep{zhao2024vidit} employs fine-grained grouping with dynamic quantization.
Preliminary progress has also been made in quantizing video generation models~\citep{feng2025q, li2025dvd}.
However, these methods still degrade significantly under extreme low-bit settings, motivating us to explore the underlying challenges and propose LRQ-DiT.
In this work we focus on diffusion transformers and propose a low-bit weight–activation quantization framework tailored for DiTs to enhance the quality of image and video generation.

%% file: Sec/3_preliminaries.tex
\section{Preliminaries}
\subsection{Quantization}
The uniform quantizer maps floating-point data $\bm{x}$ to integer data $\bm{x}_q$, with a given bitwidth $b$. The quantization and dequantization process can be expressed as:
\begin{equation}
\begin{aligned}
\label{eq: Initialize model2}
\bm{x}_{q} &= \text{clamp}\left(\left\lfloor\frac{\bm{x}}{s}\right\rceil - z, 0, 2^{b}-1\right), \\
{\bm{x}_{f}} &= s \cdot (\bm{x}_{q} + z) \approx \bm{x}, \\
\text{where}~{s} &= \frac{\bm{x}_{max} - \bm{x}_{min}}{2^b-1}, {z} = \left\lfloor \frac{\bm{x}_{min}}{s} \right\rceil.
\end{aligned}
\end{equation}
Among them, $\left\lfloor\cdot\right\rceil$ is the rounding function, \text{clamp}$(\cdot)$ is the function that limits the value to the range of $[0, 2^{b}-1]$, $s$ denotes the scaling factor, and $z$ represents the zero-point.

\subsection{Rotation Transformation}
Recently, some rotation-based methods~\citep{ashkboos2024quarot,liu2024spinquant,lin2024duquant} eliminate outliers in LLM activations via the rotation matrix $\mathbf{H}$ along with corresponding weight transformations. 
This transformation improves the quantization performance of the model while ensuring computational invariance.
\begin{equation}
\begin{aligned}
\mathbf{Y} &= \mathbf{X} \mathbf{W}^{\top} = (\mathbf{X} \mathbf{H}) (\mathbf{W} \mathbf{H})^{\top}, \\
\text{where} &~\mathbf{H} \mathbf{H}^{\top} = I \quad \text{and} \quad | \mathbf{H}| = 1.
\end{aligned}
\end{equation}
$\mathbf{H}$ is typically a Hadamard matrix, which is orthogonal and has elements taking values in $\{+1, -1\}$. A Hadamard matrix is a square matrix of size $2^n$, as shown in the Equation~(\ref{eq:hadamard}).
\begin{equation}
\begin{aligned}
\mathbf{H}_2 = \frac{1}{\sqrt{2}} 
\begin{bmatrix}
1 & 1 \\
1 & -1
\end{bmatrix}
\quad \text{and} \quad
\mathbf{H}_{2^n} = \mathbf{H}_2 \otimes \mathbf{H}_{2^{n-1}}.
\label{eq:hadamard}
\end{aligned}
\end{equation}

%% file: Sec/4_method.tex
\section{Method}
\subsection{Twin-Log Quantization for Weights}
\begin{figure}[!t]
\centering
\includegraphics[width=0.8\linewidth]{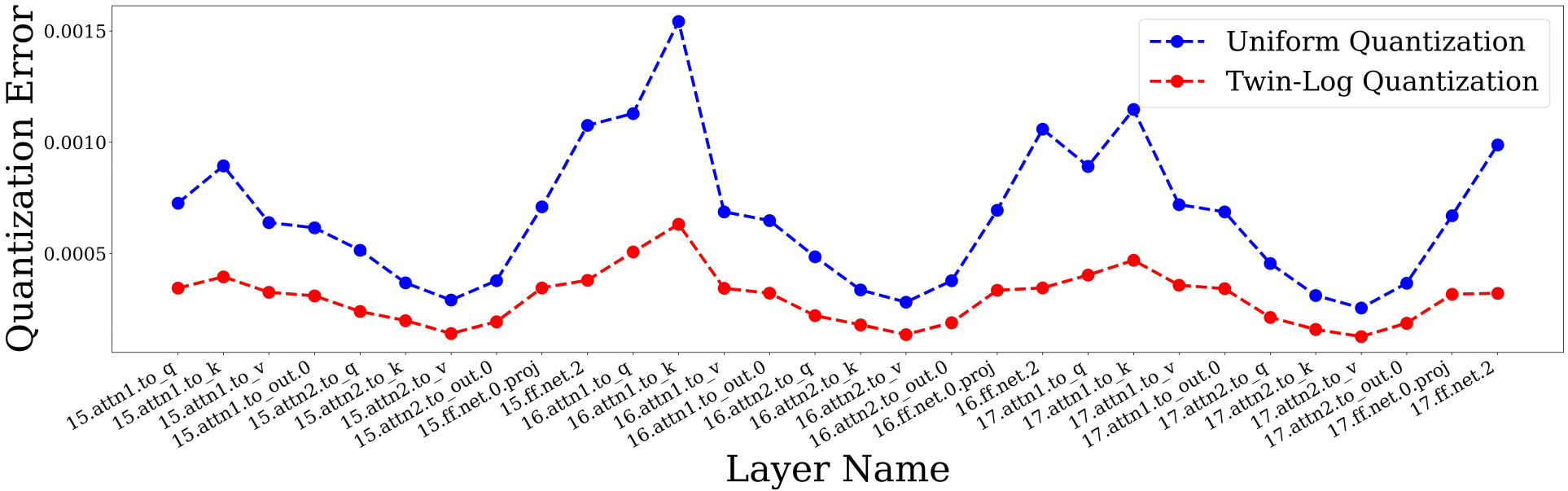} 
\caption{Comparison of 3-bit quantization errors of PixArt weights ($||W_q-W||_2$)  under different quantization methods.}
\vskip -0.1in
\label{fig:quantization error}
\end{figure}

\textbf{OB 1.} 
Uniform quantization mismatches the Gaussian distribution of weights, leading to significant quantization errors and limiting the performance of low-bit quantized models.
\paragraph{Motivation}
Inspired by previous studies~\citep{li2025dvd, feng2025q} and our experiments, we visualize the weight distributions of DiTs and observe that most weight matrices follow an approximately zero-mean Gaussian-like distribution with long tails on both sides, as shown in Figure~\ref{fig:overview}(a).
Existing works~\citep{zhao2024vidit,wu2024ptq4dit} primarily adopt uniform quantizers for weight quantization.
However, our preliminary experiments reveal that 3-bit uniform quantization introduces a large error ($||W_q-W||_2$) on PixArt weights, illustrated in Figure~\ref{fig:quantization error}.
This explains the severe accuracy degradation under low-bit settings.
We attribute this to two factors:
\ding{202} uniform quantizer cannot allocate sufficient intervals to the dense central region, limiting the expressive power of the majority of parameters; 
and \ding{203} uniform quantizer fails to handle the long-tail regions, where parameters are few but have large magnitudes.

\paragraph{Twin-Log Quantization (TLQ)}
To address the limitations of uniform quantization, we propose a twin-log quantization (TLQ) method designed to better fit the Gaussian-like weight distribution. 
Regarding issue \ding{202}, we introduce a log-based quantizer to allocate more quantization intervals to the dense central region. 
Specifically, we first apply a logarithmic transformation to obtain $\bm{W}' = \log_2(|\bm{W}|)$. 
The transformed weight matrix $\bm{W}'$ is then divided into positive and negative parts using position masks $\bm{M^{+}}$ and $\bm{M^{-}}$: 
\begin{equation}
\begin{aligned}
\bm{W}^{+} = \bm{W}' \cdot \bm{M^{+}}, \quad 
\bm{W}^{-} = \bm{W}' \cdot \bm{M^{-}} .
\end{aligned}
\end{equation}
where $\bm{M^{+}}$ and $\bm{M^{-}}$ indicate the positions of positive and negative values in $\bm{W}$, respectively. 
We quantize the positive and negative parts separately, and the $b$-bit quantization is finally performed as:
\begin{equation}
\begin{aligned}
\label{eq: log-based quantizer}
\bm{W}_{q}^{+} &= \text{clamp}\left(\left\lfloor\frac{\bm{W}^{+}}{s^{+}}\right\rceil - z^{+}, 0, 2^{b-1}-1\right), \\
\bm{W}_{q}^{-} &= \text{clamp}\left(\left\lfloor\frac{\bm{W}^{-}}{s^{-}}\right\rceil - z^{-}, 0, 2^{b-1}\right),  \\
\bm{W}_{q}^{~} ~&= \bm{W}_{q}^{+}  -  \bm{W}_{q}^{-}  .
\end{aligned}
\end{equation}
where $s^{+}$ and $s^{-}$ denote the scaling factors, and $z^{+}$ and $z^{-}$ are zero-points for the positive and negative parts, respectively. 
This log-based design also naturally de-emphasizes the long-tail regions of the weight distribution, where parameters are few but have large magnitudes. 
As shown in Figure~\ref{fig:overview}(a), this simultaneously addresses issue~\ding{203}.

\paragraph{Search-based Clipping}  
Inspired by learnable weight clipping (LWC)~\citep{shao2023omniquant}, we adopt a search-based clipping strategy to further suppress quantization errors in long-tail regions. 
Specifically, two hyperparameters, $\alpha$ and $\beta$, are introduced to clip the extreme values on both sides of the weight distribution, and their optimal values are determined via a simple grid search.  
The clipping-based scaling factors and zero-points are calculated as:  
\begin{equation}
\begin{aligned}
\label{eq: hyperparameters clip}
s^{+} &= \frac{\alpha \cdot \bm{W}_{\max}^{+} - \bm{W}_{\min}^{+}}{2^{b-1}-1}, 
\quad z^{+} = \left\lfloor \frac{\bm{W}_{\min}^{+}}{s^{+}} \right\rceil,\\
s^{-} &= \frac{\beta \cdot \bm{W}_{\max}^{-} - \bm{W}_{\min}^{-}}{2^{b-1}}, 
\quad z^{-} = \left\lfloor \frac{\bm{W}_{\min}^{-}}{s^{-}} \right\rceil.
\end{aligned}
\end{equation}
The de-quantized weights are reconstructed as:
\begin{equation}
\begin{aligned}
\label{eq: dequant}
\bm{W}_{f} = 
2^{\,s^{+}(\bm{W}_{q}^{+} + z^{+})} \cdot \bm{M}^{+}
- 2^{\,s^{-}(\bm{W}_{q}^{-} + z^{-})} \cdot \bm{M}^{-}.
\end{aligned}
\end{equation}
This clipping strategy effectively suppresses extreme long-tail weights while preserving the dense regions of the distribution, further improving low-bit quantization performance.

\paragraph{Hardward Implementation}
Since the exponential operations in Equation~(\ref{eq: dequant}) are not integer-friendly, existing TLQ still has potential for further acceleration in practical deployment. 
To address this, we design a hardware-oriented acceleration pipeline by decomposing the exponential term into an integer part and a residual part. 
Specifically, for the quantized weights, we split the exponents into:
\begin{equation}
\begin{aligned}
\label{eq: integer and residual}
f^{+} &= \left\lfloor s^{+}(\bm{W}_{q}^{+}+z^{+}) \right\rfloor, 
\quad r^{+} = s^{+}(\bm{W}_{q}^{+}+z^{+}) - f^{+}, \\
f^{-} &= \left\lfloor s^{-}(\bm{W}_{q}^{-}+z^{-}) \right\rfloor, 
\quad r^{-} = s^{-}(\bm{W}_{q}^{-}+z^{-}) - f^{-}, \\
\bm{W}_{f} &= 2^{f^{+}} \cdot 2^{r^{+}} \cdot \bm{M}^{+} 
- 2^{f^{-}} \cdot 2^{r^{-}} \cdot \bm{M}^{-}.
\end{aligned}
\end{equation}
We then introduce an integerization factor $2^{-I}$ (e.g., 1/64, 1/128, 1/256)~\citep{wu2024adalog} to approximate the residual exponents by integers while minimizing quantization error:
\begin{equation}
\begin{aligned}
\label{eq: adalog and integer}
2^{r^{+}} &\approx \mathbb{I}^{r+} \cdot 2^{-I} = 
\left\lfloor \frac{2^{r^{+}}}{2^{-I}} \right\rceil \cdot 2^{-I}, \\
2^{r^{-}} &\approx \mathbb{I}^{r-} \cdot 2^{-I} = 
\left\lfloor \frac{2^{r^{-}}}{2^{-I}} \right\rceil \cdot 2^{-I}.
\end{aligned}
\end{equation}
Combining Equations~(\ref{eq: integer and residual}) and~(\ref{eq: adalog and integer}), the weight–activation multiplication can be expressed as:
\begin{equation}
\begin{aligned}
\bm{W}\bm{X} &=
~\left(2^{f^{+}} \mathbb{I}^{r+}\bm{M}^{+} 
- 2^{f^{-}} \mathbb{I}^{r-}\bm{M}^{-}\right) 2^{-I} \mathbb{I}^{X}  s^{X} \\
&=~~~ [(\mathbb{I}^{r+}\mathbb{I}^{X}) << (f^{+}-I)]  \bm{M}^{+}  s^{X} \\
& \quad - [(\mathbb{I}^{r-}\mathbb{I}^{X}) << (f^{-}-I)] \bm{M}^{-}  s^{X}.
\end{aligned}
\end{equation}
$\mathbb{I}^{X}$ and $ s^{X}$ represent the integers and scales of activations.
This formulation enables efficient hardware execution: the SHIFT operation accelerates power-of-two multiplications, and mask multiplications are implemented via fast bitwise AND operations.  
Therefore, TLQ can be deployed with lower latency and reduced computational cost, while maintaining high quantization accuracy.
We provide the speedup measurement in Section~\ref{subsec_speed}.

\subsection{Adaptive Rotation Scheme for Activations}
\textbf{OB 2.} 
We identify two types of activation outliers in DiTs: Mild Outliers with slightly elevated values and Salient Outliers with larger magnitudes in specific channels.
\paragraph{Motivation}
We observe two distinct types of activation outliers in DiT models. Here, outliers refer to values significantly larger than typical activations. 
Based on their magnitude, we categorize them as ~\ding{202} \textbf{Mild Outliers} and ~\ding{203} \textbf{Salient Outliers}. 
As shown in Figure \ref{fig:overview}(b), most outliers slightly exceed typical activation values and \textit{rarely surpass 5}, which we term Mild Outliers. 
In contrast, Salient Outliers appear in specific channels with \textit{much larger magnitudes}.
For instance, in the blocks.26.attn2.to$\_$q layer of PixArt, there is a notable phenomenon of numerical outliers: 3.6\% of the activation values exceed 10, while a small but critical subset (0.1\%) surpasses 100, with the peak value reaching as high as 245.
These disproportionately large activations pose significant challenges for low-bit weight-activation quantization.
Besides, these Salient Outliers are prompt-independent and consistently observed across models such as PixArt and FLUX, with more pronounced patterns in PixArt. 

Notably, our observation differs from the Massive Outliers reported in LLMs~\citep{sun2024massive,liu2024intactkv}, which typically appear in a few specific tokens. In contrast, both Mild and Salient Outliers in DiT models are distributed across all tokens but confined to specific channels.

\paragraph{Adaptive Rotation Scheme (ARS)}
Recently, rotation-based transformations have been proven effective in mitigating outliers in LLMs. 
The use of Hadamard matrices $\mathbf{R}_{H} \in \{+1, -1\}^{n \times n}$ provides a simple and efficient way to smooth outliers~\citep{ashkboos2024quarot,chee2023quip,tseng2024quip}.
Moreover, DuQuant~\citep{lin2024duquant} further enhances performance on massive outliers by constructing rotation matrices based on the largest outlier channel and introducing permutation transformations to balance the activation distribution. 
The Hadamard rotation is lightweight and straightforward, while DuQuant’s operations are more effective for handling high-magnitude outliers.

For DiT models, where Mild and Salient Outliers appear in different linear layers, we propose an adaptive rotation scheme tailored to different activation inputs. Specifically, we introduce an adaptive metric $\mathbf{J}$ to measure the fluctuation degree in layer activations with shape $BNC$:
\begin{equation}
\begin{aligned}
\label{eq: QuIP J}
\mathbf{J} &= \frac{\| \mathbf{X} \|_F}{\sqrt{B  N  C}}   \\ 
\end{aligned}
\end{equation}
Here, $\| \mathbf{X} \|_F$ captures the overall activations' magnitude, and the term ${\sqrt{BNC}}$ normalizes the scale across layers.
Based on this metric, we apply Hadamard rotation to activation matrices with $\mathbf{J}$ lower than a predefined threshold, while activations with larger $\mathbf{J}$ adopt DuQuant’s dual transformations.
\begin{equation}
\begin{aligned}
\mathbf{X} &= 
\begin{cases} 
\mathbf{X} \cdot \mathbf{R}_{H} & \text{if } \mathbf{J} < \mathbf{Threshold}, \\
\mathbf{X} \cdot \hat{\mathbf{R}}_{1} \cdot \mathbf{P} \cdot \hat{\mathbf{R}}_{2}  & \text{if } \mathbf{J} \geq \mathbf{Threshold}.
\end{cases}
\end{aligned}
\end{equation}
\input{Tables/table_pixart_sigma}

We follow DuQuant to construct an approximate rotation matrix $\hat{\mathbf{R}}$ for suppressing channel-wise outliers. 
First, the channel with the largest outlier concentration, $c^*$, is identified. 
A rotation matrix is greedily updated by swapping $c^*$ to the first channel and applying an orthogonal rotation $\mathbf{Q}$:
\begin{equation}
\mathbf{R}^i = \mathbf{E}_{c^*}\,\tilde{\mathbf{R}}\,\mathbf{Q}\,\mathbf{E}_{c^*}, 
\quad 
\hat{\mathbf{R}}=\prod_{i=1}^{n}\mathbf{R}^i
\end{equation}
where $\mathbf{E}_{c^*}$ is a switch matrix exchanging the first and $c^*$ columns, and $\tilde{\mathbf{R}}$ is an initialized orthogonal matrix. 
We perform $n$ greedy updates on the top-$k$ influential channels to obtain $\hat{\mathbf{R}}_{1}$, which mitigates outliers more effectively than a fixed Hadamard matrix.
To further balance inter-block outlier distributions, channels are reordered in a staggered manner by sorting their numerical ranges and alternately assigning them across blocks, denoted as an orthogonal matrix $\mathbf{P}$.
After permutation, another rotation $\hat{\mathbf{R}}_{2}$ is applied to further smooth residual outliers, effectively improving low-bit quantization performance.
Our adaptive rotation scheme effectively mitigates both types of outliers while balancing computational cost and quantization performance. 
The adaptive rotation scheme is particularly effective for DiT models, as it handles the two types of outliers in different ways to balance efficiency and performance. 
Notably, the entire strategy is training-free, avoiding expensive optimization for hundreds of linear layers in DiT models.
Using the calibration data, we apply additional rotations to about 5\% of layers that exhibit Salient Outliers. 
Simply setting the threshold to 1 yields satisfactory results in practice. 
The resulting computational overhead during inference is negligible compared to the performance benefits achieved.
We can achieve satisfactory results by simply setting the threshold to 1. With the help of calibration data, we only perform additional processing on 5\% of the layers that contain salient outliers. Compared to the benefits gained, the overhead introduced during the inference phase is negligible.

\input{Tables/table_pixart_alpha}

\subsection{Theoretical Analysis}
\begin{lemma}
Let $x\in \mathbb{R}^C$ be an activation vector and $R\in\mathbb{R}^{C\times C}$ be an orthogonal rotation matrix. The transformed vector $x'=xR$ preserves the L2-norm:
\begin{equation*}
    \|x'\|_2=\|x\|_2.
\end{equation*}
Furthermore, if $x$ is dominated by a salient outlier such that its L-infinity norm is $\|x\|_{\infty}$, the transformed vector's L-infinity norm is suppressed as:
\begin{equation*}
    \|x'\|_{\infty}\approx \frac{\|x\|_{\infty}}{\sqrt{C}}
\end{equation*}
\end{lemma}

\begin{theorem}
\label{theorem1}
Let $X$ be a C-dim activation tensor containing a salient outlier of magnitude $\alpha=\|X\|_{\infty}$. Let $\textrm{MSE}_{\textrm{before}}=\textrm{MSE}(\textrm{TLQ}(X))$ and $\textrm{MSE}_{\textrm{after}}=\textrm{MSE}(\textrm{TLQ}(\textrm{ARS}(X)))$ denote the mean squared error of Twin-Log Quantization before and after the application of ARS, respectively. Then, the ratio of these errors is given by:
$$
\frac{\textrm{MSE}_{\textrm{after}}}{\textrm{MSE}_{\textrm{before}}}\approx (1-\frac{log_2\sqrt{C}}{log_2\alpha})^2.
$$
\end{theorem}

\begin{proof}
We first map the input $x$ into the logarithmic domain via $y=log_2(|x|)$, where uniform quantization with step size $\Delta_y$ introduces an error $e_y$. Mapping back to the linear domain, the quantized value is $x_q=|x|\cdot 2^{e_y}$, leading to a linear-domain error,
$$
e_x=x_q-|x|=|x|(2^{e_y}-1)\approx |x|(ln2)e_y,
$$
where the approximation follows from the first-order expansion of $2^{e_y}$. Hence the mean squared error is,
$$
\textrm{MSE}_x= (|x|ln2)^2\sigma_{e_y}^2,
$$
showing the quantization error scales quadratically with the input amplitude and grows rapidly in the presence of outliers.

The step size $\Delta_y$ is determined by the dynamic range in the log domain, which is dominated by the maximum absolute value: $\Delta_y\propto log_2(||X||_{\infty})$. If $\|X\|_{\infty}=\alpha$, then $\Delta_y\propto log_2(\alpha)$. After applying \textit{ARS}, the maximum is suppressed to $\|X'\|_{\infty}\approx \alpha/\sqrt{C}$, leading to a reduced step size $\Delta'_y\propto log_2(\alpha/\sqrt{C})$. Since \textit{ARS} is an orthogonal transform, the Frobenius norm is preserved, i.e. $\|X'\|_F^2=\|X\|_F^2$.

Therefore, the total MSE before and after \textit{ARS} are,
$$
\textrm{MSE}_{\textrm{before}} \propto (log_2\alpha)^2\|X\|_F^2, \textrm{MSE}_{\textrm{after}} \propto (log_2\frac{\alpha}{\sqrt{C}})^2\|X\|_F^2,
$$
and their ratio simplifies to,
$$
\frac{\textrm{MSE}_{\textrm{after}}}{\textrm{MSE}_{\textrm{before}}}\approx (1-\frac{log_2\sqrt{C}}{log_2\alpha})^2,
$$
which quantifies the error reduction due to the shrinkage of the effective dynamic range by \textit{ARS}.
\end{proof}

This result quantitatively demonstrates the powerful synergy between \textit{ARS} and \textit{TLQ}. It reveals that the error reduction becomes more pronounced as the vector dimensionality, C, increases. \textit{ARS} leverages high dimensionality to effectively eliminate large outliers, reshaping the data into a more compact, low-magnitude region. By removing the sparse, high-magnitude tail, \textit{ARS} allows \textit{TLQ}'s precision-allocation mechanism to operate with maximum efficiency, leading to a substantial and scalable reduction in quantization error.

%% file: Tables/table_pixart_sigma.tex
\begin{table*}[!t]
    \fontsize{9}{\normalbaselineskip}\selectfont 
    \setlength{\tabcolsep}{1mm}
    \small 
    \caption{
    Performance of LRQ-DiT (PixArt-$\Sigma$) text-to-image generation on different prompt sets.
    FID is calculated based on the images generated by FP16 inference.
    To ensure fairness in the comparison, we align the experimental setup and rerun SmoothQuant, QuaRot, DuQuant, ViDiT-Q, denoted as \textbf{Method*}.
    We follow the setting of ViDiT-Q to preserve FP precision for certain sensitive and parameter small layers (e.g., embed, norm$\_$out, proj$\_$out, adaln$\_$single, caption$\_$projection), denoted as [\textbf{WnAn}]\textsuperscript{\textdagger}. These settings also apply to other Tables.
    }
    \vskip 0.1in
    \label{tab:pixart_sigma}
    \centering
    \resizebox{1\linewidth}{!}{ 
    \begin{tabular}{ccccccc|cccc|cccc}
    \toprule
    \multirow{2}{*}{Model} &   \multirow{2}{*}{Precision} & \multirow{2}{*}{Method}& \multicolumn{4}{c}{COCO prompt set} & \multicolumn{4}{c}{MJHQ prompt set} & \multicolumn{4}{c}{sDCI prompt set} \\
    \cmidrule(lr){4-7} \cmidrule(lr){8-11} \cmidrule(lr){12-15}
         & & & FID $\downarrow$ & IR $\uparrow$ & SSIM $\uparrow$ & PSNR $\uparrow$ & FID $\downarrow$ & IR $\uparrow$ & SSIM $\uparrow$ & PSNR $\uparrow$ & FID $\downarrow$ & IR $\uparrow$ & SSIM $\uparrow$ & PSNR $\uparrow$\\
    \midrule
     &\multirow{5}{*}{W3A4\textsuperscript{\textdagger}}  &  Smooth*  &437.47  &-2.28 &0.08 &~6.92 &444.72 &-2.28 &0.09 &~6.74 &429.29 &-2.28 &0.09 &~6.97 \\
     & &  QuaRot*  &274.75 &-2.03 &0.36 &10.28 &223.76 &-1.72 &0.40 &11.14 &261.14 &-1.94 &0.31 &10.71 \\
     & &  DuQuant*  &251.61 &-1.59 &0.31 &10.66 &143.33 &-0.59 &0.37 &11.96 &190.79 &-1.26 &0.33 &11.69\\
     & &  ViDiT-Q*   &210.55 &-1.33 &0.35 &10.77 &200.83 &-1.53 &0.36 &11.01 &239.69 &-1.66 &0.30 &10.62 \\
     & &  Ours & \textbf{~90.15} & \textbf{~0.43} & \textbf{0.52} & \textbf{13.47} & \textbf{~72.29} & \textbf{~0.86} & \textbf{0.52} & \textbf{14.02} & \textbf{~89.06} & \textbf{~0.47} & \textbf{0.43} & \textbf{13.46}   \\
    \cmidrule{2-15}
      &\multirow{5}{*}{W3A6\textsuperscript{\textdagger}} &  Smooth* & 496.73 & -2.28 & 0.09 & ~7.31 & 539.20 & -2.28 & 0.09 & ~7.42 & 505.05 & -2.28 & 0.11 & ~7.29 \\
      & &  QuaRot* & 152.06 & -1.22 & 0.46 & 10.87 & 107.67 & -0.39 & 0.49 &  11.73 & 134.60 & -1.09 & 0.40 & 11.30 \\
      & &  DuQuant*  &122.92 &-0.53 &0.45 &11.25 &~94.74 &~0.15 &0.33 &10.13 &~95.61 &-0.31 &0.26 &9.84\\
      & &  ViDiT-Q* & 115.63 & -0.30 & 0.42 & 11.41 & ~99.05 & -0.22 & 0.46 & 11.86 & 124.75 & -0.65 & 0.37 & 11.54 \\
     & &  Ours & \textbf{~52.97} & \textbf{~0.79} & \textbf{0.53} & \textbf{13.47} &  \textbf{~47.37} &  \textbf{~1.10} & \textbf{0.54} &  \textbf{14.16} &  \textbf{~53.61} &  \textbf{~0.68} &  \textbf{0.45} &  \textbf{13.24} \\
    \cmidrule{2-15}
       &\multirow{5}{*}{W3A8\textsuperscript{\textdagger}} &  Smooth* & 501.06 & -2.28 & 0.09 & ~7.25 & 569.98 & -2.28 & 0.09 & ~7.35 & 510.94 & -2.28 & 0.11 & ~7.31 \\
      &  &  QuaRot* & 144.83 & -1.15 & 0.47 & 10.89 & 103.68 & -0.32 & 0.49 & 11.75 & 129.51 &  -1.04 & 0.40 & 11.33 \\
      & &  DuQuant*  &119.01 &-0.37 &0.41 &11.68 &~88.47 &~0.31 &0.30 &10.03 &~91.00 &-0.26 &0.37 &11.36\\
      & &  ViDiT-Q* & 111.21 & -0.25 & 0.43 & 11.44 & ~96.71 & -0.17 & 0.46 & 11.88 & 120.07 & -0.59 & 0.38 & 11.65 \\
       \textbf{PixArt-$\Sigma$}  &  &  Ours & \textbf{~51.06} &  \textbf{~0.82} & \textbf{0.52} & \textbf{13.36} &  \textbf{~47.56} &  \textbf{~1.12} & \textbf{0.58} &  \textbf{14.51} &  \textbf{~53.56} &  \textbf{~0.80} &  \textbf{0.45} &  \textbf{13.51} \\
      \cmidrule{2-15}
      \textbf{20 Steps}   &\multirow{5}{*}{W4A4\textsuperscript{\textdagger}}  &  Smooth* & 392.64 & -2.28 & 0.16 & ~7.01 & 397.50 & -2.28 & 0.13 & ~6.25 & 389.65 & -2.28 & 0.15 & ~6.80 \\
      & &  QuaRot* & ~86.03 & ~0.65 & 0.47 & 13.01 & ~67.24 & ~1.02 & 0.51 & 14.17 & ~81.64 & ~0.62 & 0.40 & 13.05 \\
      & &  DuQuant*  &~74.70 &~0.73 &0.48 &13.64  & ~57.09 &~1.16 &0.55 &15.31  &~75.54 &~0.72 &0.22 &9.66\\
      & &  ViDiT-Q* & ~67.65 & ~0.78 & 0.53 & 14.40 & ~60.08 & ~1.12 & 0.35 & 10.97 & ~68.61 & ~0.81 & 0.25 & 10.05 \\
      & &  Ours & \textbf{~65.63} &  \textbf{~0.83} & \textbf{0.54} & \textbf{14.47} &  \textbf{~53.87} &  \textbf{~1.16} & \textbf{0.55} &  \textbf{15.44} &  \textbf{~61.94} &  \textbf{~0.82} &  \textbf{0.43} &  \textbf{13.77} \\
    \cmidrule{2-15}
      &\multirow{5}{*}{W4A6\textsuperscript{\textdagger}}  &  Smooth*  & ~97.31 & ~0.20 &0.35 &11.24 &~82.86 &~0.42 &0.35 &11.26 &~94.96 &-0.09 &0.32 &10.90  \\
    & &  QuaRot*   &~42.96 &~0.86 &0.56 &13.91 &~38.97 &~1.21 &0.60 &15.05 &~43.02 &~0.93 &0.49 &13.95  \\
      & &DuQuant* &~40.43 &~0.91 &0.59 &14.44 &~42.35 &~1.26 &0.35 &10.68 &~39.66 &~0.99 &0.53 &14.87  \\
      & &  ViDiT-Q*   &~38.08 &~0.90 &0.61 &15.14 &~34.99 &~1.25 &0.64 &16.21 &~38.12 &~0.97 &0.55 &15.05 \\
      & &  Ours   &\textbf{~35.83} &\textbf{~0.95} &\textbf{0.63} &\textbf{15.56} &\textbf{~32.79} &\textbf{~1.27} &\textbf{0.66} &\textbf{16.66} &\textbf{~35.57} &\textbf{~1.00} &\textbf{0.57} &\textbf{15.69}  \\
        \cmidrule{2-15}
      &\multirow{5}{*}{W4A8\textsuperscript{\textdagger}}  &  Smooth*  &~72.51 &~0.48 &0.36 &11.33 &~67.76 &~0.68 &0.36 &11.32 & ~72.14 &~0.23 &0.33 &11.06   \\
    & &  QuaRot*   &~41.78 &~0.83 &0.56 &13.94 &~37.90 &~1.21 &0.61 &15.08 &~41.80 &~0.93 &0.50 &14.01   \\
      & &  DuQuant*  &~40.17 &~0.90 &0.58 &14.40 &~34.98 &~1.24 &0.63 &15.96 &~38.41 &~0.98 & 0.54 &14.96 \\
      & &  ViDiT-Q*  &~36.96 &~0.88 &0.62 &15.17 &~34.23 &~1.24 &0.65 &16.25 &~37.41 &~0.98 &0.56 &15.13   \\
      & &  Ours   &\textbf{~35.16} &\textbf{~0.95} &\textbf{0.64} &\textbf{15.64} &\textbf{~32.39} &\textbf{~1.27} &\textbf{0.66} &\textbf{16.53} &\textbf{~34.14} &\textbf{~0.99} &\textbf{0.59} &\textbf{15.82}   \\ 
    \bottomrule
    \end{tabular}
    }
\end{table*}

%% file: Tables/table_pixart_alpha.tex
\begin{table*}[!t]
    \fontsize{9}{\normalbaselineskip}\selectfont 
    \setlength{\tabcolsep}{1mm}
    \small 
    \caption{
    Performance of LRQ-DiT (PixArt-$\alpha$) text-to-image generation on different prompt sets.
    }
    \vskip 0.1in
    \label{tab:pixart_alpha}
    \centering
    \resizebox{1\linewidth}{!}{ 
    \begin{tabular}{ccccccc|cccc|cccc}
    \toprule
    \multirow{2}{*}{Model} &   \multirow{2}{*}{Precision} & \multirow{2}{*}{Method}& \multicolumn{4}{c}{COCO prompt set} & \multicolumn{4}{c}{MJHQ prompt set} & \multicolumn{4}{c}{sDCI prompt set} \\
    \cmidrule(lr){4-7} \cmidrule(lr){8-11} \cmidrule(lr){12-15}
         & & & FID $\downarrow$ & IR $\uparrow$ & SSIM $\uparrow$ & PSNR $\uparrow$ & FID $\downarrow$ & IR $\uparrow$ & SSIM $\uparrow$ & PSNR $\uparrow$ & FID $\downarrow$ & IR $\uparrow$ & SSIM $\uparrow$ & PSNR $\uparrow$\\
    \midrule
     &\multirow{5}{*}{W3A4\textsuperscript{\textdagger}}  &  Smooth*  &321.59 &-2.26 &0.40 &~7.28 &350.92 &-2.27 &0.35 &~4.53 &329.29 &-2.26 &0.39 &~8.22 \\
     & &  QuaRot*  &292.01 &-2.25 &0.36 &10.93 &280.91 &-2.19 &0.36 &11.89 &271.63 &-2.20 &0.35 &11.27 \\
     & &  DuQuant*  &269.46 &-2.12 &0.26 &10.07 &194.97 &-1.60 &0.34 &11.47 &225.82 &-2.07 &0.30 &10.97\\
     & &  ViDiT-Q*   &245.37 &-2.04 &0.35 &10.37 &232.04 &-1.86 &0.31 &11.89 &216.56 &-2.03 &0.32 &11.61 \\
     & &  Ours & \textbf{~90.36} & \textbf{~0.01} & \textbf{0.49} & \textbf{12.82} & \textbf{~75.51} & \textbf{~0.48} & \textbf{0.54} & \textbf{13.90} & \textbf{~87.06} & \textbf{-0.25} & \textbf{0.48} & \textbf{13.22}   \\
     \cmidrule{2-15}
     &\multirow{5}{*}{W3A6\textsuperscript{\textdagger}}  &  Smooth*  & 299.82 & -2.26  & 0.40 & ~8.15 & 293.05 & -2.26 & 0.41 & ~7.07 & 327.55 & -2.27 & 0.39 & ~7.28 \\
     & &  QuaRot* & 186.82 & -2.08 & 0.43 & 11.10 & 176.33 & -1.90 & 0.46 & 12.10 & 170.67 & -2.03 & 0.42 & 11.47 \\
      & &  DuQuant* &162.18 &-1.49 &0.44 &11.38 &165.78 &-1.54 &0.45 &11.51 &163.11 &-1.68 &0.35 &11.33  \\
     & &  ViDiT-Q*  & 151.59 & -1.46 & 0.43 & 10.78 & 124.88 & -0.87 & 0.43 & 12.45 & 138.73 & -1.63 & 0.39 & 11.85  \\
     & &  Ours & \textbf{~57.76}  & \textbf{~0.31} & \textbf{0.53} & \textbf{12.86} & \textbf{~51.95} & \textbf{~0.85} & \textbf{0.56} & \textbf{14.12} & \textbf{~61.41} & \textbf{~0.03} & \textbf{0.49} & \textbf{13.36} \\
    \cmidrule{2-15}
      &\multirow{5}{*}{W3A8\textsuperscript{\textdagger}}  &  Smooth*  & 307.17 & -2.26 & 0.39 & ~7.97 & 303.78 & -2.26 & 0.41 & ~6.61 & 316.17 & -2.27 & 0.39 & ~7.34 \\
    &  &  QuaRot* & 183.38 & -2.06 & 0.43 & 11.09 & 170.95 & -1.89 & 0.47 & 12.10 & 167.23 & -2.03 & 0.42 & 11.47 \\
      & &  DuQuant*  &166.73 &-1.56 &0.45 &10.93  &126.23 &-1.01 &0.49 &12.15 &146.43 &-1.47 &0.33 &11.10\\
     & &  ViDiT-Q*  & 149.14 & -1.43 & 0.41 & 10.43 & 121.65 & -0.82 & 0.44 & 12.47 & 136.03 & -1.61 & 0.39 & 11.84  \\
    \textbf{PixArt-$\alpha$}   & &  Ours & \textbf{~56.95}  & \textbf{~0.30} & \textbf{0.52} & \textbf{12.83} & \textbf{~50.84} & \textbf{~0.87} & \textbf{0.57} & \textbf{14.15} & \textbf{~61.17} & \textbf{~0.05} & \textbf{0.49} & \textbf{13.35} \\
    \cmidrule{2-15}
       \textbf{20 Steps}  &\multirow{5}{*}{W4A4\textsuperscript{\textdagger}}  &  Smooth*  & 375.94 & -2.25 & 0.42 & ~9.85 & 367.77 & -2.26 & 0.43 & 10.09 & 385.19 & -2.26 & 0.39 & ~9.60 \\
       & &  QuaRot* & 120.81 & ~0.26 & 0.19 & ~9.86 & ~79.17 & ~0.79 & 0.33 & 12.87 & 135.87 & -0.13 & 0.25 & 11.54 \\
      & &  DuQuant*   &~71.25 &~0.61 &0.51 &14.10 &~58.42 &~1.10 &0.55 &14.81  &~75.92 &~0.49 &0.46 &14.10 \\
    &  &  ViDiT-Q*  & ~72.12 & ~0.64 & 0.49 & 13.78 & ~60.45 & ~1.12 & 0.47 & 14.59 & ~78.69 & ~0.51 & 0.41 & 13.72  \\
      & &  Ours & \textbf{~61.83}  & \textbf{~0.77} & \textbf{0.51} & \textbf{14.30} & \textbf{~50.18} & \textbf{~1.24} & \textbf{0.56} & \textbf{15.41} & \textbf{~74.33} & \textbf{~0.55} & 0.41 &13.85 \\
    \cmidrule{2-15}
      &\multirow{5}{*}{W4A6\textsuperscript{\textdagger}}  &  Smooth*  &405.53 &-2.18 &0.13 &~8.33 &336.22 &-1.97 &0.13 &~9.08 &324.68 &-1.94 &0.14 &~8.87  \\
    & &  QuaRot*   &~42.71 &~0.75 &0.58 &14.63 &~38.04 &~1.27 &0.61 &15.62 & ~45.01&~0.70 &0.54 &14.68  \\
      & &  DuQuant*   &~42.06 &~0.72 &0.60 &14.74 &~38.84 &~1.20 &0.62 &15.79 &~42.41 &~0.67 &0.56 &15.06\\
      & &  ViDiT-Q*   &~40.92 &~0.74 &0.60 &14.62 &~37.96 &~1.28 &0.62 &15.84 &~44.27 & ~0.73&0.55 &14.87  \\
      & &  Ours   &\textbf{~36.58} &\textbf{~0.75} &\textbf{0.62} &\textbf{15.47} &\textbf{~33.92} &\textbf{~1.32} &\textbf{0.65} &\textbf{16.50} &\textbf{~40.74} &\textbf{~0.77} &\textbf{0.56} &\textbf{15.50}  \\
        \cmidrule{2-15}
      &\multirow{5}{*}{W4A8\textsuperscript{\textdagger}}  &  Smooth*  &407.81 &-2.18 &0.12 &~8.32 &354.63 &-2.02 &0.13 & ~8.93&334.11 &-1.96 &0.13 &~8.71  \\
    & &  QuaRot*   & ~41.41& ~0.77& 0.59&14.67 &~37.48 &~1.27 &0.62 &15.67 &~44.11 &~0.68 &0.55 &14.73  \\
      & &  DuQuant*  &~42.04 &~0.83 &0.58 &14.70 &~37.03 &~1.29 &0.61 &15.35 &~42.11 &~0.01 &0.56 &14.82\\
      & &  ViDiT-Q*    &~40.59 &~0.73 &0.60 &14.64 &~37.90 &~1.25 &0.63 &15.87 &~42.55 &~0.73 &0.56 &14.97  \\
      & &  Ours    & \textbf{~37.06} &\textbf{~0.83} & \textbf{0.62} & \textbf{15.35} & \textbf{~33.68} & \textbf{~1.32}  & \textbf{0.65} & \textbf{16.56} & \textbf{~39.75} & \textbf{~0.78} & \textbf{0.57} & \textbf{15.59} \\
    \bottomrule
    \end{tabular}
    }
\end{table*}

%% file: Sec/5_exp.tex
\section{Experiment}
\subsection{Experimental Setting}
\paragraph{Models and Prompt sets} 
Following prior studies~\citep{zhao2024vidit, li2024svdquant}, we evaluated performance across different DiT models: PixArt-$\Sigma$ (steps=20, CFG=4.5), PixArt-$\alpha$ (steps=20, CFG=4.5), FLUX.1-schnell (steps=4, CFG=0.), and FLUX.1-dev (steps=50, CFG=3.5).
We selected 1024 prompts from COCO~\citep{chen2015microsoft}, MJHQ-30K~\citep{li2024playground}, and the summarized Densely Captioned Images (sDCI)~\citep{urbanek2024picture} respectively to evaluate the performance of LRQ-DiT in text-to-image generation tasks.
In addition, we conducted performance validation on the text-to-video generation task on the OpenSORA v1.2 model, generating 51 frames and 240p videos. 
Specifically, we use the OpenSORA prompt set (including 10 prompts) and the Vbench prompt set (including 251 prompts) for verification.

\input{Tables/table_flux}

\paragraph{Evaluation Metrics} 
We evaluated both the quality and similarity of images generated by the quantized model.
Specifically, we utilize the Fr\'echet Inception Distance (FID, the lower the better)~\citep{heusel2017gans} to measure the distributional distance between images generated by the quantized and FP16 models. 
Then, we employ Image Reward (IR, the higher the better)~\citep{xu2023imagereward} to estimate human preference ratings for the generated images.
We adopt Peak Signal Noise Ratio (PSNR, the higher the better) to measure the numerical similarity of the generated images.
Structural Similarity Index Measure (SSIM, the higher the better) is used to measure the luminance, contrast, and structure similarity of the images generated by the quantized and FP16 models. 
In addition, we use Imaging Quality and Aesthetic Quality to measure the quality of the video. 
Measure the consistency of generated videos using Background Consist, Subject Consist, Scene Consist, and Overall Consist. 
We also consider metrics such as PSNR, SSIM, VQA-Overall, Motion Smooth, Dynamic Degree, etc. for these generated videos.

\paragraph{Implementation Details and Baselines}  
We have aligned the experimental settings of ViDiT-Q~\citep{zhao2024vidit}, which include employing per-channel static quantization for weights and per-token dynamic quantization for activations, along with the migration strength of $0.5$ to accomplish difficulty transfer from activations to weights. Besides, we retain full precision for certain sensitive layers with a small number of parameters.
Our PTQ method utilizes 4-10 prompts to obtain the smoothing factor, metric $\mathbf{J}$, rotation matrix $\mathbf{R}$, and permutation matrix $\mathbf{P}$ during the calibration phase, without the need for retraining the model or fine-tuning parameters. Therefore, it does not introduce additional overhead during the inference phase.
Compared to the performance gain brought by our method, the additional computational overhead is acceptable. 
We compare LRQ-DiT with methods SmoothQuant~\citep{xiao2023smoothquant}, QuaRot~\citep{ashkboos2024quarot}, DuQuant~\citep{lin2024duquant} and the current state-of-the-art PTQ method ViDiT-Q~\citep{zhao2024vidit} to validate the effectiveness of our method. To ensure fairness in the comparison of experimental results, we rerun these methods on both PixArt, FLUX and OpenSORA models using A6000 and A100 GPUs, respectively.

\subsection{Main Results}
\paragraph{PixArt}
We have conducted validation for the text-to-image generation tasks on the COCO, MJHQ, and sDCI datasets using both PixArt-$\Sigma$ and PixArt-$\alpha$, and evaluated their performance under different settings: W3A4, W3A6, W3A8, W4A4, W4A6, and W4A8.
The experimental results in Table \ref{tab:pixart_sigma} and \ref{tab:pixart_alpha} demonstrate that text-to-image generation cannot be achieved solely through a naive smoothing scheme.
QuaRot mitigates the impact of outliers in the model, enabling the generation of moderately satisfactory images at high-bit width. However, once quantized to low-bit width, the model's performance experiences a sharp decline.
ViDiT-Q, as the current state-of-the-art PTQ method, performs well under 4-bit weight quantization, but the performance deteriorates once the weights are quantized to 3-bit.
LRQ-DiT achieves excellent performance levels under 3-bit weight quantization and further enhances the model's performance under other settings.

\input{Tables/table_opensora} 
\paragraph{FLUX}
Based on previous research findings~\citep{li2024svdquant} and our experimental observations, we have reached the following conclusion: due to FLUX inherent advantages, the models are capable of generating high-quality images even after low-bit quantization.
Specifically, only using naive smoothing scheme to achieve 4-bit quantization, the model still exhibits great performance.
We conducted 3-bit weight quantization experiments on the FLUX.1-schnell and FLUX.1-dev models and validated their performance on different datasets.
The results in Table \ref{tab:flux} demonstrate that our method enhances the model's performance under the W3A4 quantization setting, with improvements observed in both the quality and similarity of the generated images.

\paragraph{OpenSORA}
We validate the performance of the OpenSORA v1.2 model in text-to-video generation tasks on 10 prompts in the OpenSORA prompt set and 251 prompts in the Vbench prompt set.
We measure the quality, consistency, smoothness, dynamics, and other characteristics of the generated videos through various evaluation metrics.
As shown in the Table~\ref{tab:opensora_vbench}, our method exhibits strong performance advantages compared to other comparative methods~(SmoothQuant, QuaRot, DuQuant, and ViDiT-Q) in different bit settings and various prompt sets.
Especially under low-bit quantization settings (W3A4, W3A6, and W3A8), LRQ-DiT exhibits stronger robustness and improves the performance of quantized models.
\input{Tables/table_ablation}

\begin{figure}[!t]
\centering
\includegraphics[width=0.5\linewidth]{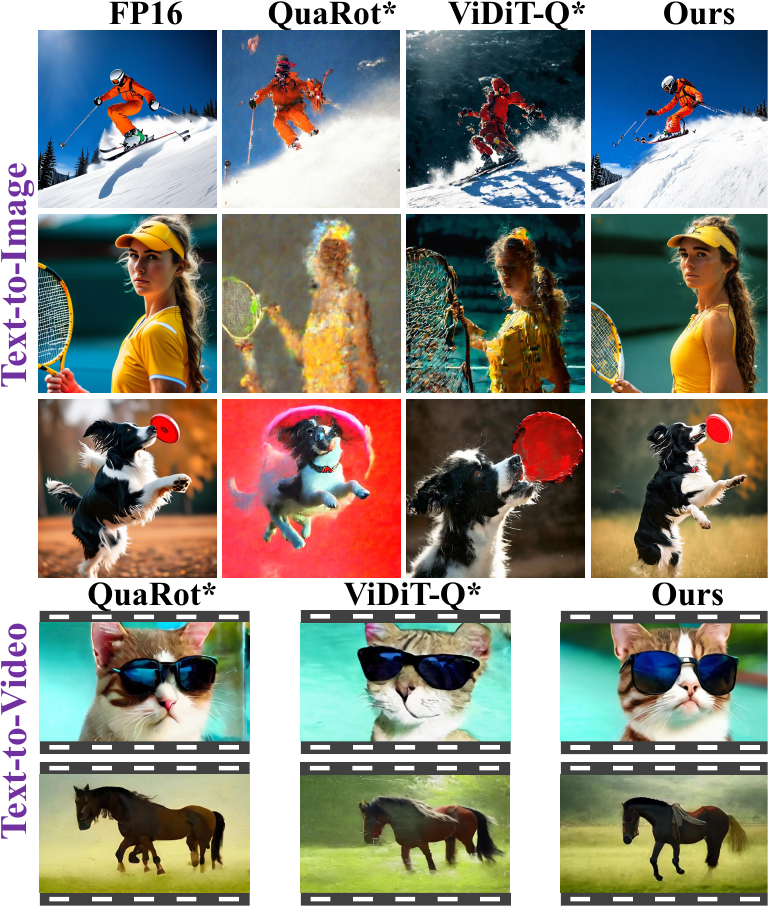} 
\vskip 0.1in
\caption{Visual Comparison of Images generated by PixArt-$\Sigma$ and Videos generated by OpenSORA using several different W3A6 quantization methods.}
\label{fig:Visualization_Image}
\vskip 0.1in
\end{figure}

\subsection{Ablation Study}
We conducted ablation experiments in Table \ref{tab:ablation} to validate the effectiveness of TLQ and ARS.
Without TLQ and ARS, the model exhibits poor performance, and images are also lacking in information.
When ARS is applied independently to address the impact of mild and salient outliers in the model, the model's performance is enhanced.
When TLQ is applied independently to achieve more precise quantization of weights, the model's performance also improves significantly.
When TLQ and ARS are applied together, the quantized model achieves optimal performance in both image quality and similarity, providing strong evidence of the synergistic effect between TLQ and ARS.
Compared to directly applying ``Hadamard" or ``DuQuant" rotation to all activations, ARS demonstrates significant advantages: it not only offers higher precision but also has quite limited computational overhead.

\subsection{Visualization Comparison}
We generated images using QuaRot, ViDiT-Q, and our method LRQ-DiT respectively on a set of prompts.
The images and videos generated by different methods are shown in the Figure \ref{fig:Visualization_Image}.
After comparison, it can be seen that our method demonstrates significant advantages in generation tasks. The foreground details in the generated images by our method are presented with clarity and specificity, the depiction of motion scenes is precise and vivid, and at the same time, character recognition becomes much easier. 
Compared with other methods, the videos generated by our method have better quality, smoother flow, and stronger consistency.

\input{Tables/table_latency}

\subsection{SpeedUp and Latency}
\label{subsec_speed}

To validate the hardware efficiency of our method LRQ-DiT, we implemented a custom Triton-based kernel for INT8 matrix multiplication. 
In our setup, both weights and activations are quantized to INT8, and the matrix multiplication is performed entirely in INT8 before accumulation into higher-precision registers. 
All experiments were conducted on a single NVIDIA RTX 4090 GPU, where for each matrix shape, we automatically selected the optimal block configuration through heuristic search to maximize hardware utilization. 
Each kernel was executed for 10 iterations with warm-up runs to eliminate startup overhead, and the averaged runtime was reported. 
The results, summarized in Table~\ref{tab:lantency}, demonstrate that INT8 execution achieves a consistent $2.15-2.45\times$ speedup over FP16 across different matrix sizes, with throughput improvements from approximately 65 TFLOPs to over 160 TFLOPs. These findings provide strong empirical evidence that the proposed quantization method not only preserves accuracy but also unlocks substantial acceleration on modern GPUs.

%% file: Tables/table_flux.tex
\begin{table*}[!t]
    \fontsize{9}{\normalbaselineskip}\selectfont 
    \setlength{\tabcolsep}{1mm}
    \small 
    \caption{
        Performance of LRQ-DiT (FLUX.1) text-to-image generation on different prompt sets.
    }
    \vskip 0.1in
    \label{tab:flux}
    \centering
    \resizebox{1\linewidth}{!}{ 
    \begin{tabular}{cc|cccccc|cccccc}
    \toprule
     \multirow{4}{*}{Precision} &\multirow{4}{*}{Method} & \multicolumn{6}{c}{\textbf{FLUX.1-schnell 4 Steps}}   & \multicolumn{6}{c}{\textbf{FLUX.1-dev 50 Steps}}  \\
     \cmidrule(lr){3-8} \cmidrule(lr){9-14}  
     & & \multicolumn{2}{c}{COCO prompt set} & \multicolumn{2}{c}{MJHQ prompt set} & \multicolumn{2}{c}{sDCI prompt set}  &  \multicolumn{2}{c}{COCO prompt set} & \multicolumn{2}{c}{MJHQ prompt set} & \multicolumn{2}{c}{sDCI prompt set} \\
    \cmidrule(lr){3-4} \cmidrule(lr){5-6} \cmidrule(lr){7-8}     \cmidrule(lr){9-10} \cmidrule(lr){11-12} \cmidrule(lr){13-14}
      &  & FID $\downarrow$ & IR $\uparrow$  & FID $\downarrow$ & IR $\uparrow$ & FID $\downarrow$ & IR $\uparrow$      & FID $\downarrow$ & IR $\uparrow$  & FID $\downarrow$ & IR $\uparrow$ & FID $\downarrow$ & IR $\uparrow$ \\
    \midrule   
   \multirow{4}{*}{W3A4\textsuperscript{\textdagger}} &  Smooth* &     71.66 & 0.48 & 78.35 & 0.36 & 72.35 & 0.19     & 156.64 & -1.23  & 146.33 & -1.15 & 166.03 & -1.72 \\
 &  QuaRot* &    60.46 & 0.85  & 65.09 & 0.84 & 63.73 & 0.68 & ~98.37  & ~0.14 & ~96.68 & ~0.31 & ~96.78 & -0.27 \\
  &  ViDiT-Q* & 59.07 & 0.87 & 63.71 & 0.87 & 62.25 & 0.73  & ~91.18  & ~0.29  & ~90.29 & ~0.47 & ~91.01 & -0.07\\
   &  Ours & \textbf{50.77} &  \textbf{0.92} &  \textbf{54.52} &  \textbf{0.98}   &  \textbf{56.40} &  \textbf{0.79}     &  \textbf{~58.81} &  \textbf{~0.57} & \textbf{~63.13} & \textbf{~0.76} &  \textbf{~63.58} &  \textbf{~0.25}  \\
    \midrule   
    \multirow{3}{*}{W3A6\textsuperscript{\textdagger}} & Smooth* &     42.61 & 0.91  & 46.80 & 1.13 & 45.32 & 0.92      & ~41.74 & ~0.85  & ~44.87 & ~1.15 & ~45.71 & ~0.81 \\
     &QuaRot* &      41.52 & 0.99  & 44.97 & 1.24 & 44.40 & 1.03     &~39.73  & ~0.94 & ~43.21 & ~1.22 & ~44.14 & ~0.94 \\
    &Ours &      \textbf{41.47} & \textbf{0.99}  & \textbf{40.81} & \textbf{1.26} & \textbf{43.74} & \textbf{1.03}     & \textbf{~38.60} &  \textbf{~0.94} & \textbf{~39.57} & \textbf{~1.22} &  \textbf{~43.31} &  \textbf{~0.95}  \\
    \midrule   
  \multirow{3}{*}{W3A8\textsuperscript{\textdagger}} &    Smooth* &    41.93 & 0.90  & 46.67 & 1.14 & 44.65 & 0.92      & ~41.06 & ~0.87  & ~43.72 & ~1.16 & ~44.85 & ~0.84 \\
    &   QuaRot*  &   \textbf{40.74} & \textbf{1.00}  & 44.62 & 1.24 & 43.68 & 1.01       & ~39.20  & ~0.94 & ~41.97 & ~1.22 & ~43.94 & ~0.93 \\
    &  Ours &   41.36 & 0.98  & \textbf{40.43} & \textbf{1.25} & \textbf{43.56} & \textbf{1.04} \  &   \textbf{~38.48} &  \textbf{~0.95} & \textbf{~39.12} & \textbf{~1.23} &  \textbf{~42.83} &  \textbf{~0.93}  \\
    \bottomrule
    \end{tabular}
    }
\end{table*}

%% file: Tables/table_opensora.tex
\begin{table*}[!t]
    \centering
    \setlength{\tabcolsep}{0.4mm}
    \caption{Performance of the OpenSORA model in Text-to-Video Generation~(51 frames, 240p) on the OpenSORA and VBench prompt sets.}
    \vskip 0.1in
    \small
    \resizebox{1\linewidth}{!}{ 
    \begin{tabular}{cccccc|cccccccc}
    \toprule
    \multirow{3}{*}{Precison}& \multirow{3}{*}{Method} &  \multicolumn{4}{c}{OpenSORA Set~(10 prompts)} &  \multicolumn{8}{c}{Vbench Set~(251 prompts)}    \\
    \cmidrule(lr){3-6} \cmidrule(lr){7-14} 
    & & \makecell{PSNR $\uparrow$}& \makecell{SSIM $\uparrow$}& \makecell{VQA~~~~ \\ Overall $\uparrow$}& \makecell{FVD~~~\\-FP16 $\downarrow$} & \makecell{Imaging~~~ \\ Quality $\uparrow$ }& \makecell{Aesthetic~~~\\Quality $\uparrow$ } & \makecell{Motion ~~~\\ Smooth $\uparrow$} & \makecell{Dynamic ~~\\ Degree $\uparrow$}  & \makecell{Background \\Consist $\uparrow$}& \makecell{Subject~~ \\Consist $\uparrow$} & \makecell{Scene~~~ \\ Consist $\uparrow$} & \makecell{Overall~~~ \\     Consist $\uparrow$} \\ 
    \midrule
    \multirow{5}{*}{W3A4\textsuperscript{\textdagger}} &Smooth* &~7.25 &0.13 &10.25 &16.16 &48.52 &28.81 &79.28 &- &90.14 &73.38 &- &~4.03 \\
    &Quarot* &17.11 &0.59 &24.92 &~0.88 &45.83 &41.51 &97.74 &34.72 &94.14 &88.19 &15.55 &22.80 \\
    &DuQuant* &17.27 &0.59 &21.09 &~1.38 &43.31 &40.51 &98.32 &29.16 &94.63 &88.06 &20.05 &21.03 \\
    &ViDiT-Q* &17.52 &0.58 &21.07 &~1.42 & 44.58 &42.36 &98.08 &33.33 &93.87 &89.08 &20.56 &22.79\\
    &Ours &\textbf{17.87} &\textbf{0.63} &\textbf{30.29} &~1.45 &\textbf{49.43} &\textbf{47.77} &\textbf{98.56} &\textbf{40.27} &\textbf{95.13} &\textbf{93.00} &\textbf{30.08} &\textbf{25.52}\\
    \cmidrule(lr){1-14}
    \multirow{5}{*}{W3A6\textsuperscript{\textdagger}} &Smooth* &14.35 &0.51 &16.48 &0.91 &36.10 &32.93 &98.47 &4.16 &96.66 &89.89 &5.74 &14.31 \\
    &Quarot* &17.67 &0.63 &33.77 &0.43 &50.08 &45.69 &98.64 &38.88 &95.77 &92.53 &26.30 &23.93 \\
    &DuQuant* &17.00 &0.58 &23.74 &0.80 &50.71 &45.45 &98.71 &43.05 &95.92 &92.24 &32.55 &23.75 \\
    &ViDiT-Q*  &17.52 &0.61 &31.02 &0.79 &49.70 &46.89 &98.53 &43.05 &95.43 &92.10 &25.79 &24.17\\
    &Ours  &\textbf{18.40} &\textbf{0.67} &\textbf{41.48} &\textbf{0.32} &\textbf{52.61} &\textbf{51.28} &\textbf{98.93} &\textbf{52.78} &96.37 &\textbf{95.31} &\textbf{32.99} &\textbf{25.85} \\
    \cmidrule(lr){1-14}
    \multirow{5}{*}{W3A8\textsuperscript{\textdagger}} &Smooth* &14.48 &0.52 &17.32 &1.00 &38.57 &34.84  &98.45  &11.11  &96.85 &90.19 &10.10 &15.61 \\
    &Quarot* &17.69 &0.63 &34.51 &0.39 &50.31 &45.83 &98.65 &40.27 &95.80 &92.71 &28.19 &24.16 \\
    &DuQuant* &17.45 &0.62 &35.87 &1.11 &49.30 &46.82  &98.73 &37.50 &95.82 &92.92 &32.04 &24.27 \\
    &ViDiT-Q* &17.49 &0.61 &33.28 &1.07 &49.75 &47.22 &98.57 &45.83 &95.42 &92.53 &29.21 &24.20\\
    &Ours &\textbf{18.63} &\textbf{0.67} &\textbf{44.17} &\textbf{0.30} &\textbf{52.69} &\textbf{52.00} &\textbf{98.92} &\textbf{48.62} &96.40 &\textbf{95.51} &\textbf{33.22} &\textbf{25.75}\\
    \cmidrule(lr){1-14}
    \multirow{5}{*}{W4A4\textsuperscript{\textdagger}} &Smooth* &~9.35 &0.18 &~8.96 &19.44 &50.20 &29.91 &79.43 &- &92.71 &82.00 &- &~4.46 \\
    &Quarot* &19.18 &0.65 &28.79 &~0.86 &50.93 &47.53 &97.96 &47.22 &95.44 &92.30 &38.22 &25.64 \\
    &DuQuant* &20.29 &0.68 &28.90 &~0.94 &53.08 &47.90 &98.30 &51.38 &95.38 &92.55 &34.01 &25.69 \\
    &ViDiT-Q* &20.19 &0.68 &28.50 &~0.92 &51.60 &47.30  &98.14 &54.16 &95.27 &92.01 &35.61 &25.84  \\
    &Ours &\textbf{20.33}  &\textbf{0.69} &28.83 &\textbf{~0.71} &51.79 &\textbf{47.95} &\textbf{98.34} &\textbf{55.56} &\textbf{95.56} &\textbf{92.87} &37.80 &\textbf{25.87} \\
    \cmidrule(lr){1-14}
    \multirow{5}{*}{W4A6\textsuperscript{\textdagger}} &Smooth* &18.33 &0.68 &36.74 &0.31 &52.97 &47.52 &98.42 &45.83 &96.11 &94.07 &32.63 &24.47 \\
    &Quarot* &20.14 &0.72 &41.92 &0.31 &54.54 &51.34 &98.62 &56.94 &96.63 &95.13 &38.88 &25.77 \\
    &DuQuant* &20.47 &0.72 &40.69 &0.39 &55.45  &50.71 &98.66 &50.00 &96.46 &94.86 &36.33 &26.14 \\
    &ViDiT-Q*  &21.61 &0.75 &38.90 &0.40 &55.57 &50.89 &98.66 &52.77 &96.52 &94.83 &36.77 &25.98 \\
    &Ours &\textbf{22.78} &\textbf{0.79} &\textbf{42.04} &0.44 &\textbf{56.57} &\textbf{52.25} &\textbf{98.85} &48.62 &\textbf{96.90} &\textbf{95.28} &\textbf{41.28} &\textbf{26.68}\\
    \bottomrule
    \end{tabular}
    \label{tab:opensora_vbench}
    }
\end{table*}

%% file: Tables/table_ablation.tex
\begin{table*}[!t]
\centering
\fontsize{9}{\normalbaselineskip}\selectfont 
\setlength{\tabcolsep}{1mm}
\small 
\caption{Ablation studies of TLQ and ARS under different quantization settings and prompt sets on PixArt-$\alpha$.}
\vskip 0.1in
\resizebox{1\linewidth}{!}{ 
\begin{tabular}{c|ccc|cc|cc|cc|cc|cc|cc}
\toprule
\multicolumn{4}{c}{{Methods}} &\multicolumn{2}{c}{\textbf{COCO~W3A4\textsuperscript{\textdagger}}}&\multicolumn{2}{c}{\textbf{COCO~W3A6\textsuperscript{\textdagger}}}&\multicolumn{2}{c}{\textbf{COCO~W3A8\textsuperscript{\textdagger}}}&\multicolumn{2}{c}{\textbf{MJHQ~W3A4\textsuperscript{\textdagger}}}&\multicolumn{2}{c}{\textbf{MJHQ~W3A6\textsuperscript{\textdagger}}}&\multicolumn{2}{c}{\textbf{MJHQ~W3A8\textsuperscript{\textdagger}}} \\
{TLQ} & {Hadamard}& {DuQuant}& {ARS} & FID $\downarrow$ & IR $\uparrow$ & FID $\downarrow$ & IR $\uparrow$ & FID $\downarrow$ & IR $\uparrow$ & FID $\downarrow$ & IR $\uparrow$ & FID $\downarrow$ & IR $\uparrow$ & FID $\downarrow$ & IR $\uparrow$  \\
\midrule 
$\times$ & $\checkmark$ &   &   &245.37 &-2.04 &151.59 &-1.46 &149.14 &-1.43 &232.04 &-1.86 & 124.88 &-0.87 &121.65 &-0.82  \\
$\times$ &   & $\checkmark$ & &269.46 &-2.12 &162.18 &-1.49 &166.63 &-1.56 &194.97 &-1.60 &165.78 &-1.54 &126.23 &-1.01\\
$\times$ &  &  & $\checkmark$  &237.15 &-1.88 &147.70 &-1.19 &140.53 &-1.10 &212.84 &-1.65 &120.88 &-0.83 &117.77 &-0.83  \\
$\checkmark$ & $\checkmark$ &  &   &~99.50 &-0.30 &~65.61 &~0.01 &~64.87 &~0.01 &~76.02 &~0.38 &~55.14 &~0.61 &~54.85 &~0.62  \\
$\checkmark$ &   & $\checkmark$ & &~99.47 &-0.29 &~62.08 &~0.18 &~66.86 &-0.19 &~76.10 &~0.15 &~60.39 &~0.47 &~57.53 &~0.66\\
 $\checkmark$ &  &  & $\checkmark$   & \textbf{~90.36} & \textbf{~0.01} & \textbf{~57.76} & \textbf{~0.31} & \textbf{~56.95} & \textbf{~0.30} & \textbf{~75.51} & \textbf{~0.48} & \textbf{~51.95} & \textbf{~0.85} & \textbf{~50.84} & ~0.87 \\
\bottomrule
\end{tabular}
}
\label{tab:ablation}
\end{table*}

%% file: Tables/table_latency.tex
\begin{table}[!t]
\centering
\fontsize{9}{\normalbaselineskip}\selectfont 
\setlength{\tabcolsep}{0.8mm}
\caption{Performance Comparison between FP16 and INT8 GEMM (Triton Kernel). }
\vskip 0.1in
\resizebox{0.55\linewidth}{!}{ 
\begin{tabular}{cccccc}
\toprule
Matrix Shape & Block Shape & Bit & Time/ms & TFLOPs & SpeedUp \\
\midrule
\multirow{2}{*}{(2048,2048,2048)} & \multirow{2}{*}{(64,64,64)} 
& FP16 & ~0.290 & ~59.283 & ${1.00\times}$ \\
  & & INT8 & ~0.135 & 127.705 & $\mathbf{2.15\times}$ \\
\midrule
 \multirow{2}{*}{(4096,4096,4096)} &  \multirow{2}{*}{(128,128,64)}
& FP16 & ~2.102 & ~65.380 & ${1.00\times}$ \\
 & & INT8 & ~0.858 & 160.164 & $\mathbf{2.45\times}$ \\
\midrule
\multirow{2}{*}{(8192,8192,8192)} & \multirow{2}{*}{(64,256,32)} 
& FP16 & 16.810 & ~65.409 & ${1.00\times}$ \\
  & & INT8 & ~7.559 & 145.466 & $\mathbf{2.22\times}$ \\
\bottomrule
\label{tab:lantency}
\end{tabular}
}
\end{table}

%% file: Sec/6_conclusion.tex
\section{Conclusion}
This paper introduces an efficient and accurate post-training quantization method LRQ-DiT for DiT models.
LRQ-DiT identifies two key issues in model quantization and proposes corresponding solutions to address them.
Specifically, Twin-Log Quantization~(TLQ) utilizes log-based quantizer to allocate more quantization intervals to dense regions in weights, thereby reducing quantization errors and achieving precise quantization.
Then, Adaptive Rotation Scheme~(ARS) identifies mild and salient outliers in activations through the metric and suppresses the influence of different types of outliers using different rotation transformation.
We conduct detailed experiments with various bit settings on PixArt and FLUX, and then validate the performance of text-to-image generation tasks on the COCO, MJHQ, and sDCI datasets.
Besides, we also validate the performance of text-to-video generation on different prompt sets using the OpenSORA.
The experimental results demonstrate that LRQ-DiT further improves the performance of the quantized models on different bit settings and prompt sets, especially in low-bit scenarios.

%% file: Sec/7_appendix.tex
\appendix  
\section{Visual Comparison}

\begin{figure*}[!h]
\centering
\includegraphics[width=1.0\linewidth]{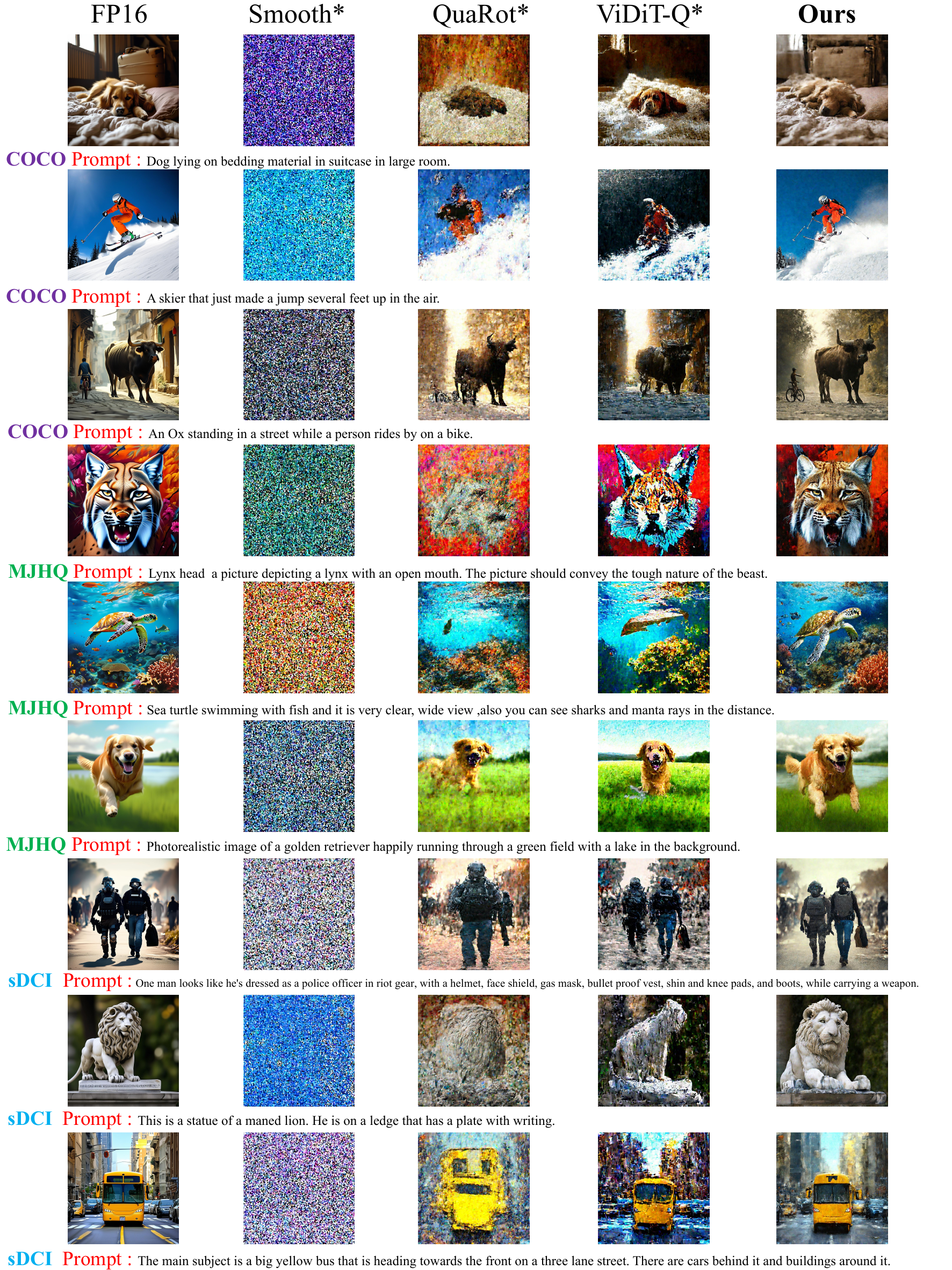} 
\caption{Visual Comparison of Generated Images from PixArt-$\Sigma$ under W3A4 Quantization.}
\label{fig:PixArt-Sigma under W3A4}
\end{figure*}
\begin{figure*}[!t]
\centering
\includegraphics[width=1\linewidth]{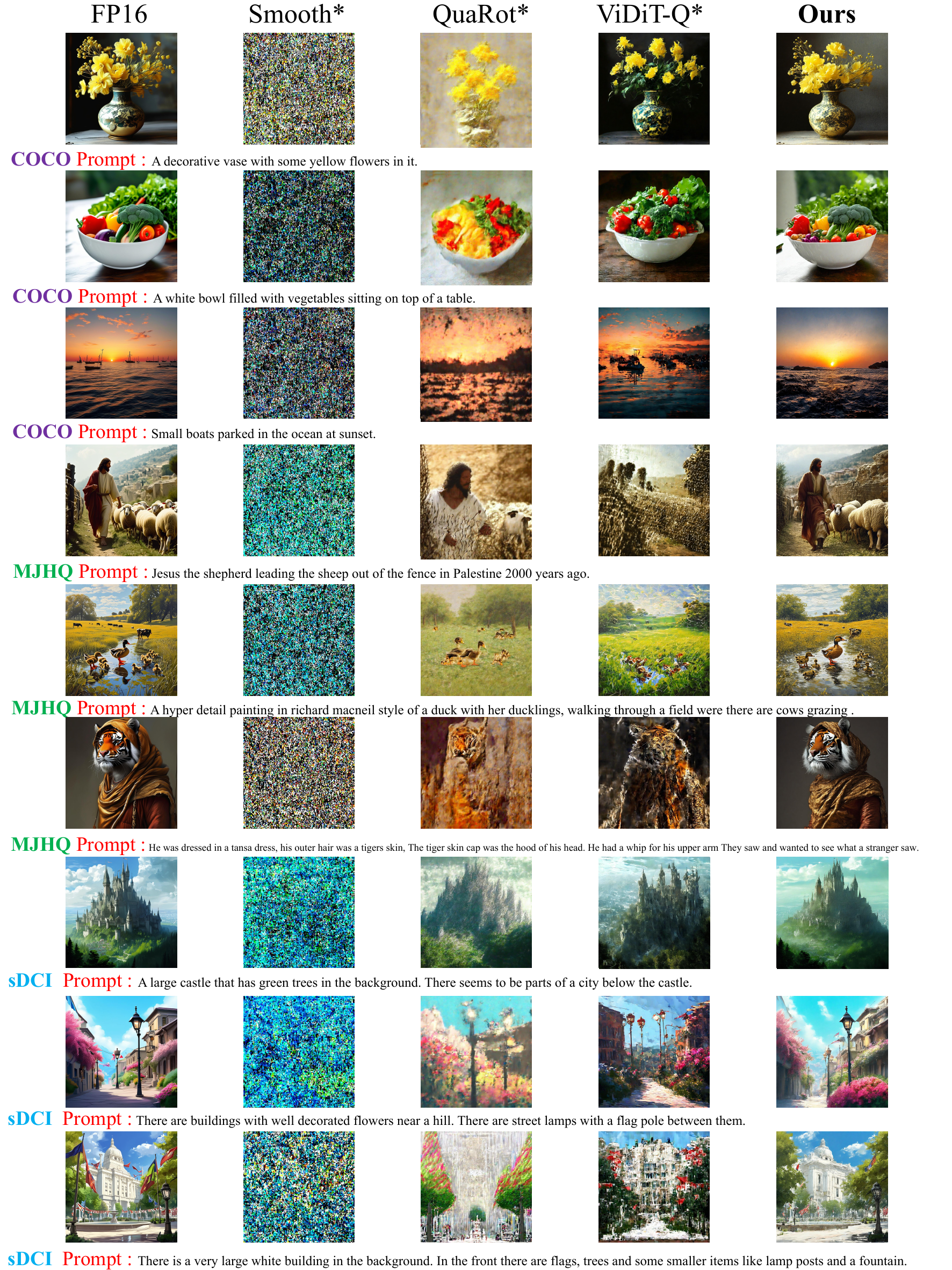} 
\caption{Visual Comparison of Generated Images from PixArt-$\Sigma$ under W3A6 Quantization.}
\label{fig:PixArt-Sigma under W3A6}
\end{figure*}
\begin{figure*}[!t]
\centering
\includegraphics[width=1\linewidth]{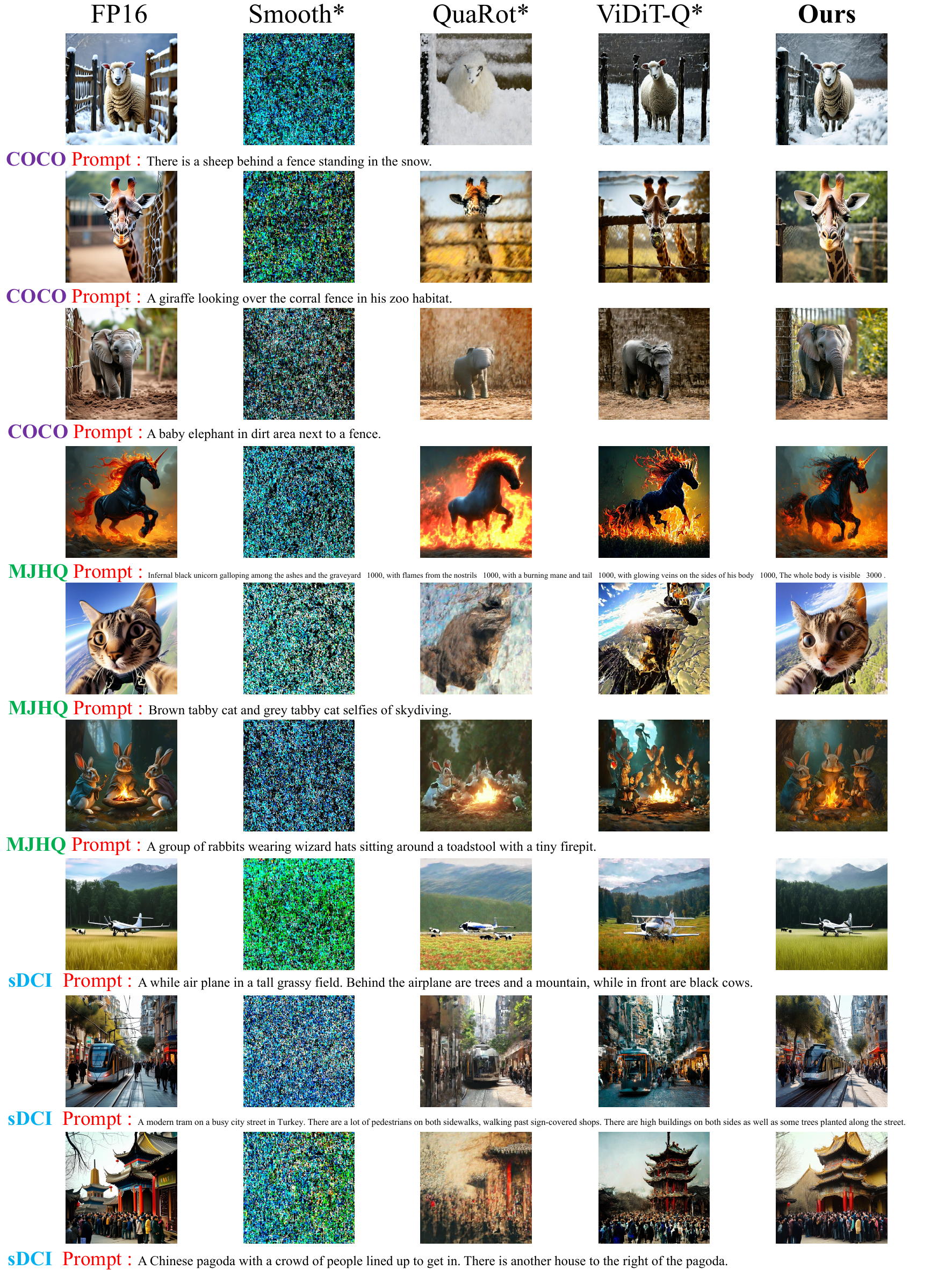} 
\caption{Visual Comparison of Generated Images from PixArt-$\Sigma$ under W3A8 Quantization.}
\label{fig:PixArt-Sigma under W3A8}
\end{figure*}
\begin{figure*}[!t]
\centering
\includegraphics[width=1\linewidth]{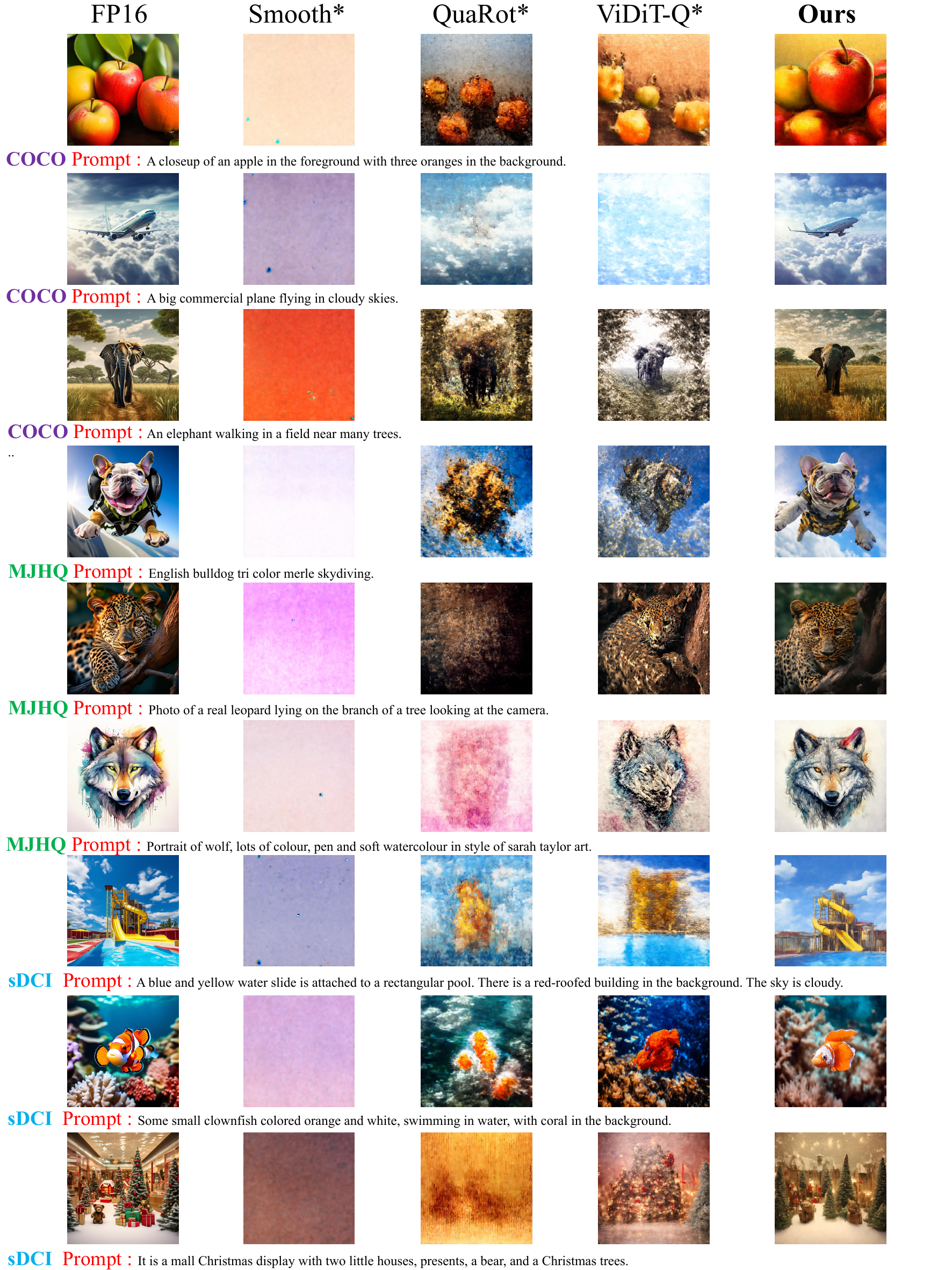} 
\caption{Visual Comparison of Generated Images from PixArt-$\alpha$ under W3A4 Quantization.}
\label{fig:PixArt-alpha under W3A4}
\end{figure*}
\begin{figure*}[!t]
\centering
\includegraphics[width=1\linewidth]{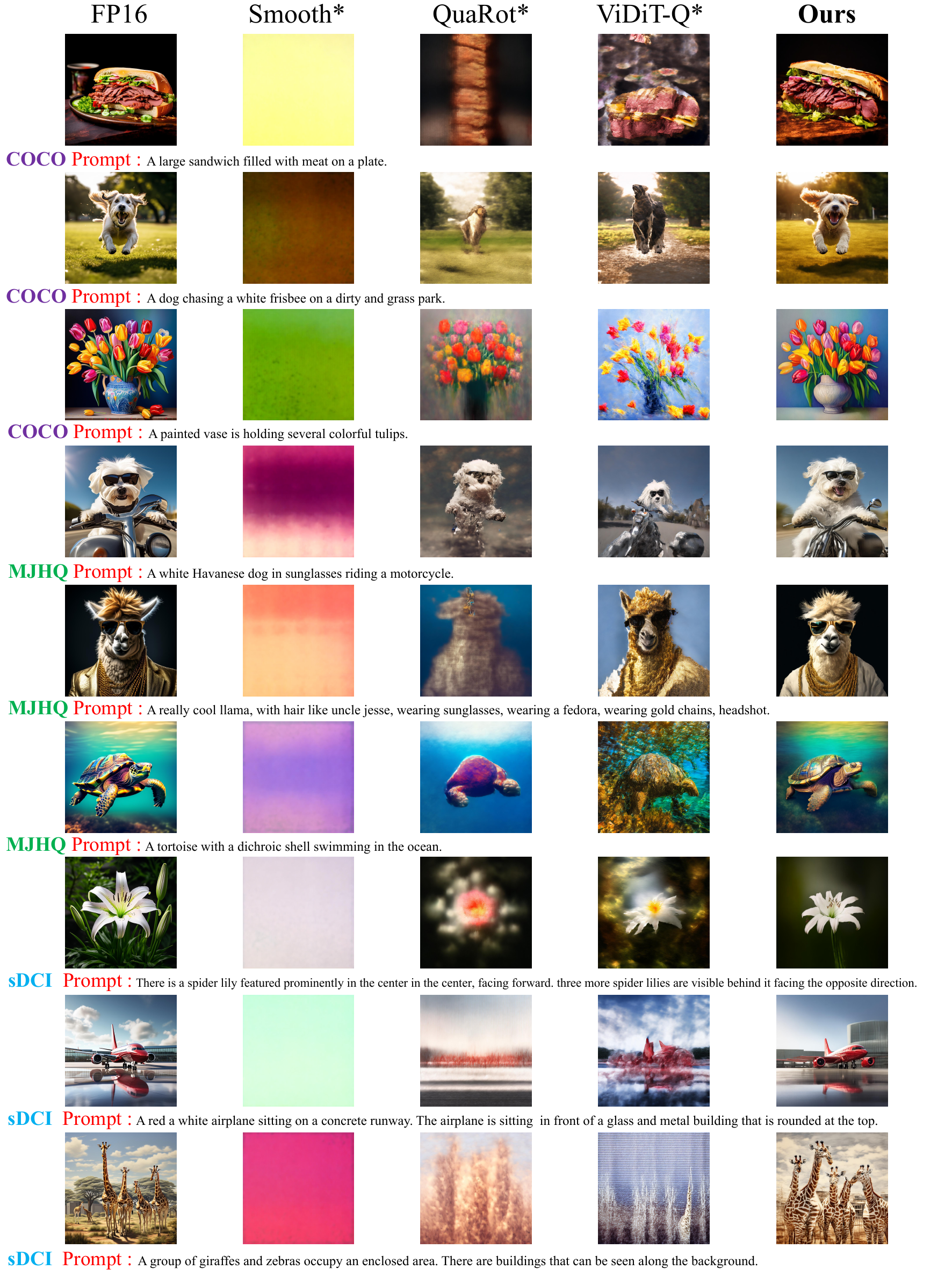} 
\caption{Visual Comparison of Generated Images from PixArt-$\alpha$ under W3A6 Quantization.}
\label{fig:PixArt-alpha under W3A6}
\end{figure*}
\begin{figure*}[!t]
\centering
\includegraphics[width=1\linewidth]{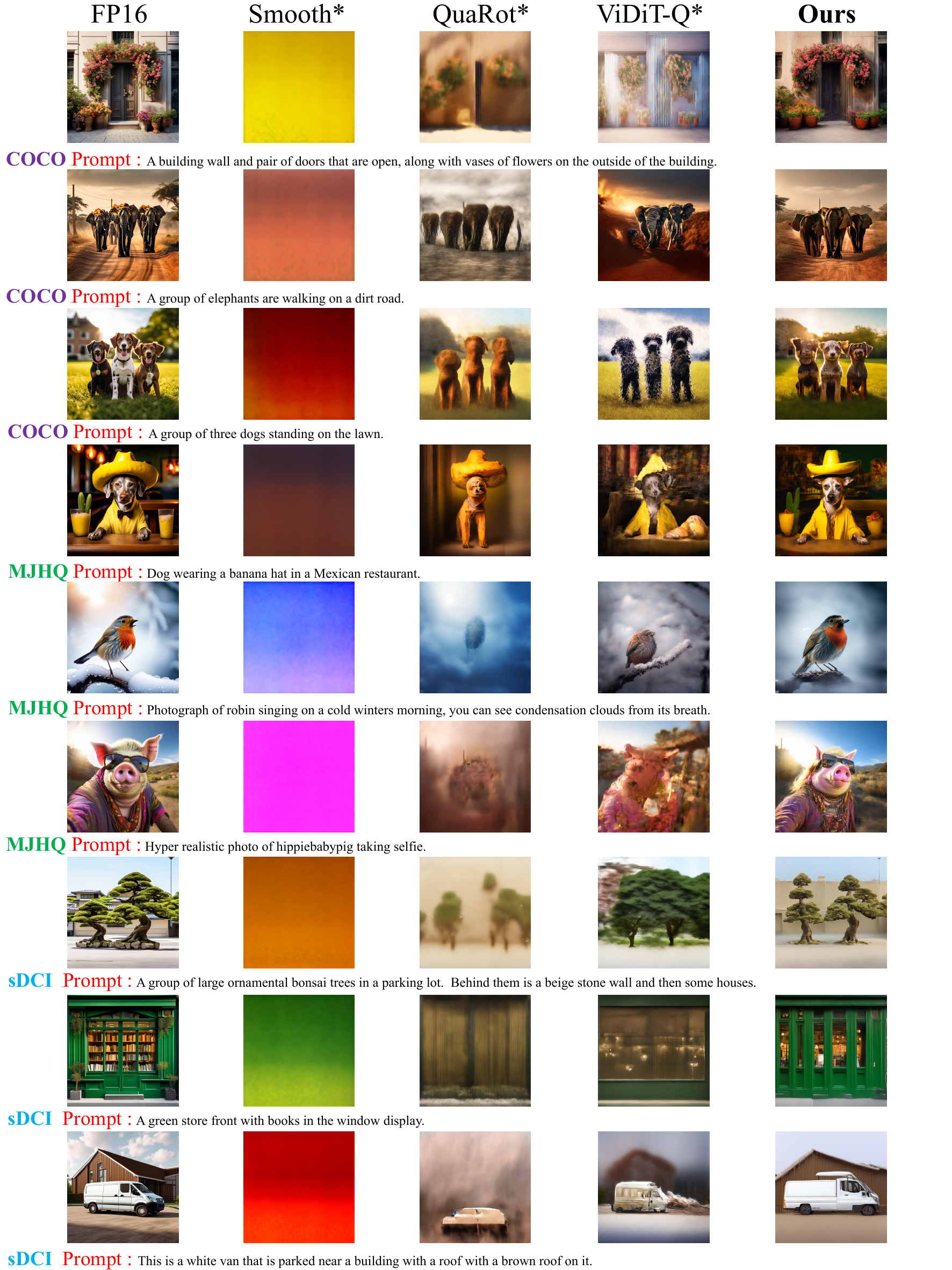} 
\caption{Visual Comparison of Generated Images from PixArt-$\alpha$ under W3A8 Quantization.}
\label{fig:PixArt-alpha under W3A8}
\end{figure*}

We provide more visual comparison results, as follows:

\begin{itemize}
\item Figure~\ref{fig:PixArt-Sigma under W3A4} shows PixArt-$\Sigma$ under W3A4 quantization.
\item Figure~\ref{fig:PixArt-Sigma under W3A6} shows PixArt-$\Sigma$ under W3A6 Quantization.
\item Figure~\ref{fig:PixArt-Sigma under W3A8} shows PixArt-$\Sigma$ under W3A8 Quantization.
\item Figure~\ref{fig:PixArt-alpha under W3A4} shows PixArt-$\alpha$ under W3A4 Quantization.
\item Figure~\ref{fig:PixArt-alpha under W3A6} shows PixArt-$\alpha$ under W3A6 Quantization.
\item Figure~\ref{fig:PixArt-alpha under W3A8} shows PixArt-$\alpha$ under W3A8 Quantization.
\end{itemize}

